\documentclass{article}
\usepackage[utf8]{inputenc}
\usepackage{lscape}
\usepackage{mathpazo}
\usepackage[margin=1.0in]{geometry} 
\usepackage{graphicx}
\usepackage{natbib}
\usepackage[labelfont=bf]{caption}

\usepackage[capposition=top]{floatrow}
\usepackage[T1]{fontenc}
\usepackage{tikz}
\usepackage{authblk}
\usepackage{xcolor}
\usepackage{booktabs}
\usepackage{multirow}

\usepackage{amsmath,amsfonts,amssymb,bm,xcolor}
\usepackage{stmaryrd}

\usepackage{booktabs} 
\usepackage{nicefrac}  
\usepackage{microtype}
\usepackage[utf8]{inputenc}

\usepackage{wrapfig}
\usepackage{multirow}

\usepackage{amssymb}
\usepackage{amsthm}
\usepackage{bbm}
\usepackage{bbm}
\usepackage{array}
\usepackage[colorlinks=true,urlcolor=purple,
linkcolor=purple,citecolor=purple]{hyperref}
\usepackage{amsmath}
\usepackage{mathtools}
\usepackage{float}

\usepackage{lscape}
\usepackage{graphicx}

\usepackage{booktabs}
\usepackage{longtable}
\usepackage{threeparttable}
\usepackage{threeparttablex}

\usepackage{mathrsfs}
\usepackage{graphicx} 
\usepackage{color}
\usepackage{natbib}

\usepackage{algorithm}
\usepackage{algpseudocode}
\usepackage{thmtools,thm-restate} 

\makeatletter
\def\paragraph{\@startsection{paragraph}{4}%
  \z@\z@{-\fontdimen2\font}%
  {\normalfont\bfseries}}
\makeatother

\usepackage{tikz}
\usetikzlibrary{decorations.markings, backgrounds, calc, positioning, shapes, shadows, arrows, fit, automata}
\usepackage{tikz-cd}
\usetikzlibrary{arrows.meta}
\tikzset{>={Latex}}

\newcounter{desccount}

\newcommand{\descref}[1]{\hyperref[#1]{#1}}

\theoremstyle{definition}

\makeatletter
\newcommand{\pushright}[1]{\ifmeasuring@#1\else\omit\hfill$\displaystyle#1$\fi\ignorespaces}
\newcommand{\pushleft}[1]{\ifmeasuring@#1\else\omit$\displaystyle#1$\hfill\fi\ignorespaces}
\makeatother

\usepackage{titlesec}
\titleformat*{\section}{\normalsize\bfseries}
\titleformat*{\subsection}{\normalsize\bfseries}
\titleformat*{\subsubsection}{\normalsize\bfseries}
\allowdisplaybreaks

\title{SciTBERT: A family of chronologically consistent language models for scientific and technological language processing}
\author{Thomas Gebhart}
\author{Russell J. Funk\thanks{E-mail correspondence to gebhart@umn.edu. Authors thank the Alfred P. Sloan Foundation (G-2024-25123) and Wellcome Leap for financial support of work related to this project.}}
\affil{{\small{Carlson School of Management, University of Minnesota}}}
\date{}

\begin{document}

\maketitle

\begin{abstract}
Pre-trained transformer models are increasingly being used to study scientific and technological progress. Encoders tuned to paper or patent text outperform general-purpose models on downstream classification, regression, and proximity tasks within science and technology. However, the applicability of these models for studying time-dependent or archival properties of science, technology, and their interface is limited due to lookahead and domain biases inherent to these pre-trained models. These limitations arise from training on corpora with unconstrained chronological and text source distributions. We introduce SciTBERT: a family of chronologically consistent BERT-derived language models trained on text from scientific papers, patents, and high-quality educational web text with training data cutoff dates spanning each year between 2013 and 2025. We also post-train these models in a chronologically-consistent manner using paper and patent citations, creating SciTBERT-CI model family. We find that these models generally outperform predecessor domain-specific encoder models even when training data is limited by early year restrictions in the corpus. To further investigate the extent to which this class of models can learn representations that bridge science and technology, we introduce the PatRepEval benchmark, a suite of patent-related text embedding tasks at the science-technology interface. Performance in a variety of classification, regression, and retrieval tasks spanning papers and patents highlights the importance of aligning encoder model representations with the domain distributions of their downstream tasks, and chronologically consistent encoders can match or exceed models trained without temporal constraints. 
\end{abstract}

\section{Introduction}\label{sec:intro}

The application of natural language processing techniques to study the structure and trajectory of scientific and technological development has undergone a profound shift in the past decades as the computational methods supporting these analyses have rapidly expanded in capability. Large language models are exemplary of this shift: generative models are now used to summarize the literature, propose hypotheses, review manuscripts, and carry out parts of the research process itself~\citep{zhang2025exploring}. Many analyses of science and technology, however, benefit from a fixed vector representation of each document rather than generated text. For literature search or retrieval, for example, these embeddings must be computed for tens or hundreds of millions of papers and patents~\citep{priem2022openalex, kinney2023semantic, hook2018dimensions}. For these uses, encoder models remain the tractable choice. An encoder maps each document to a single embedding a single forward pass, and those embeddings can be used as features in downstream classification, retrieval, and regression models. Domain-specific encoders derived from BERT~\citep{devlin2019bert} have become the standard tools for these tasks. SciBERT and BERT for Patents pre-train on scientific and patent text, respectively~\citep{beltagy2019scibert, srebrovic2020leveraging}, and SPECTER, SPECTER2, SciNCL, and PaECTER further post-train these encoders on citation links so that documents related through citations receive similar embeddings~\citep{cohan2020specter, singh2023scirepeval, ostendorff2022neighborhood, ghosh2024paecter}. These models have achieved state-of-the-art performance across classification, regression, and proximity tasks over papers or patents independently~\citep{singh2023scirepeval, ghosh2024paecter}.

Despite their strong empirical performance, these domain-specific transformer models are ill-suited for rigorous historical, predictive, and policy-oriented studies of science and technology. Existing scientific and patent language models are trained on corpora pooled across wide and often weakly documented time periods. As a result, a representation assigned to a document published in year $y$ may be generated by a model whose parameters, tokenizer, or contrastive training data encode linguistic, scientific, and technological developments that occurred only after $y$. This form of temporal leakage is analogous to the lookahead bias familiar in forecasting, finance, and retrospective policy evaluation: information unavailable at the time of the focal observation can enter the feature construction process and inflate downstream performance or distort inferred policies~\citep{kapoor2022leakage, sarkar2025lookahead, he2025chronologically, yan2026datedgpt}. The problem is especially consequential for science of science and innovation research, where text representations of papers within large language models are increasingly being used to perform and evaluate science itself. Although measures, reviews, or discoveries derived from text representations may provide contemporaneous value, the longevity of this value as science and technology progress is indeterminate, and a bias towards prior text distributions could be harmful. Similarly, because they have been trained using knowledge from the present, the validity of these representations as they are employed to understand the past becomes suspect as particular ideas or associations between concepts may have been unintelligible during the period under study. 

Chronological inconsistency can enter these models through several channels. First, subword tokenizers trained on the full corpus can encode future lexical regularities even before model pre-training begins, assigning more compact or stable tokenizations to terms that only become common past the date under study. Second, pre-training on chronologically inconsistent scientific and technological text exposes a model to future terminology and later-ingrained semantic associations between tokens that may have only emerged past some date $y$. Third, citation- or proximity-based contrastive objectives can use forward citations or topical neighborhoods that were not observable at the time of paper publication or patent filing. These sources of leakage are not only relevant for time-sensitive predictive tasks like citation prediction or renewal classification. They also affect descriptive social scientific measurement, because embeddings trained with future information may mask or distort previously salient but transient conceptual associations. This can cause historical concepts to appear more coherent, proximate, or technologically salient than they would have appeared to researchers, firms, funders, or policymakers at that time. This motivates the need for chronologically consistent representation models for scientific and technological language: models whose training corpora, relational supervision, preprocessing, and tokenization respect the temporal boundary of the downstream analysis.

To address these limitations, we introduce SciTBERT---a family of BERT-derived encoder models trained with annual chronological cutoffs over a joint corpus of scientific papers, patents, and high-quality educational web text---to address this gap. By placing papers and patents in a shared embedding space, SciTBERT supports proximity comparisons across scientific and technological documents without requiring post hoc model alignment. We enforce annual pre-training cutoff dates and chronologically consistent tokenization, enabling point-in-time representations suitable for rolling prediction, event-study designs, historical measurement, policy evaluation, and studies of scientific or technological emergence. We also introduce a family of citation-informed models, SciTBERT-CI, which are post-trained to draw together the representations of documents that are close in a chronologically truncated citation graph. To better measure technological representations and model performance at the science-technology interface, we introduce PatRepEval, a new patent and paper representation benchmarking suite. We find that these chronologically consistent models are competitive with, and often outperform, similarly-sized models containing lookahead biases across both PatRepEval and SciRepEval~\citep{singh2023scirepeval} benchmarks. 

\begin{figure}
    \centering
    \includegraphics[width=0.80\linewidth]{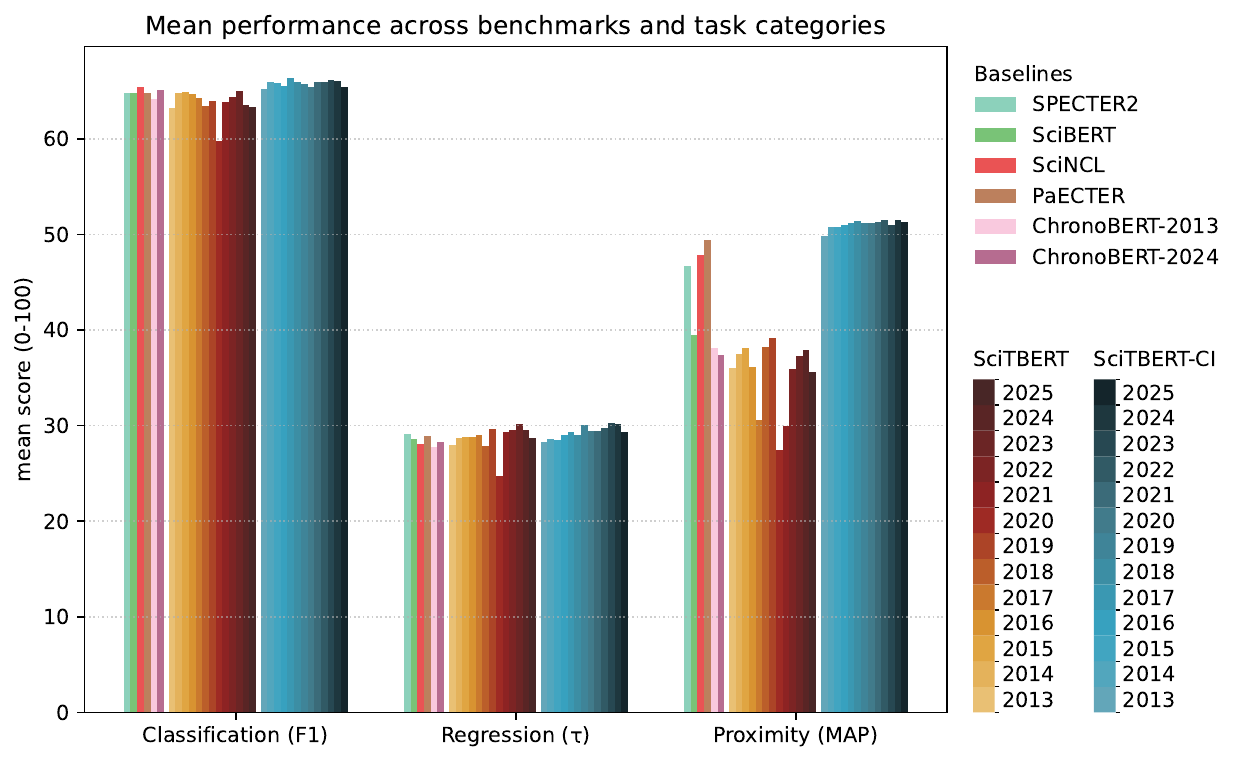}
    \caption{Performance of SciTBERT and SciTBERT-CI models, alongside a number of prior encoder models, across both SciRepEval and PatRepEval benchmarks. Performance is aggregated by taking the mean over the tasks of each category within each benchmark, then the mean of the two benchmarks. See Table~\ref{tab:results_mean_both}.}
    \label{fig:per_family_mean_both_benchmarks}
\end{figure}

\section{Background}
Early text-based approaches in science of science, innovation studies, and science policy primarily represented scientific content through keywords and their patterns of co-occurrences. For example, \citet{callon1983translations} suggested the co-occurrence of keywords could provide a constructive basis for mapping the translational behavior of research fields. Researchers used these methods to map the development of biotechnology~\citep{rip1984co}, to trace relationships between basic and technological research in polymer chemistry~\citep{callon1991co}, or to characterize the temporal evolution of scientific specialties~\citep{braam1991mapping}. A natural extension is to represent keyword counts as vectors via bag-of-words (BoW) or term-frequency inverse-document frequency (TF-IDF) and compare documents using vector similarity. Such representations have been used to map and cluster scientific disciplines~\citep{glenisson2005combining, boyack2011clustering} and to measure technological proximity among patents \citep{arts2018text, arts2023position, kelly2021measuring}. Despite their prior utility, these sparse keyword and vectorization methods are brittle, ignore word order, and struggle to capture synonymy and other nuanced semantic relationships. 

Topic modeling approaches derived from methods like Latent Dirichlet Allocation addressed this fragility somewhat by modeling documents as distributions over a smaller number of topics. \citet{blei2007correlated} applied a correlated topic model to study approximately $16,000$ articles published in \textit{Science} between 1990 and 1999. \citet{gerow2018measuring} introduced a dynamic topic modeling approach which allowed the authors to study the temporal effects of topical influence and highly-cited and prize-winning outcomes. While principled and statistically tractable, the expensive optimization procedures required to train these models do not scale to the tens- or hundreds- of millions of documents available in modern archival data sources~\citep{priem2022openalex, kinney2023semantic, hook2018dimensions}. 

The emergence of Word2Vec~\citep{mikolov2013distributed} and GloVe~\citep{pennington2014glove} introduced the ability to efficiently construct dense word vectors that better capture distributional semantics. By training shallow neural networks that force representations of words or documents appearing in similar contexts to receive similar vectorized representations, this text embedding procedure enables algebraic operations over these semantically dense latent spaces. These algebraic operations saw great success in the materials science literature~\citep{tshitoyan2019unsupervised} to recommend novel functional materials in an unsupervised manner. \citet{shibayama2021measuring} derived a measure of novelty from the average of document word embeddings and \citet{shi2023surprising} extended this framework to support multi-modal inputs. Averaging over words in a document or, more commonly, treating documents as inputs to a content prediction as in Doc2Vec~\citep{le2014distributed} allowed researchers to encode entire sentences, abstracts, or documents in the same semantically rich embedding space~\citep{jeon2022doc2vec, aceves2024mobilizing}. However, these distributed representations assign a single vector per word, regardless of context, making them ill-suited to polysemous scientific terminology. Moreover, for diachronic analysis they require separate models trained on temporal slices, requiring additional alignment procedures (e.g., Procrustes rotation~\citep{hamilton2016diachronic}) to compare embedding spaces through time. 

Transformer models address some of these issues by embedding sentences or longer documents from the contextualized sequence of their (subword) tokens derived from a multi-layer attention encoder. \citet{beltagy2019scibert} pre-trained their SciBERT model on 1.14 million scientific publications using the same training objective and architecture as the original BERT~\citep{devlin2019bert}. The authors showed that this domain specificity improved the performance of the resulting representations across a number of biomedical, computer science, and general scientific tasks. \citet{srebrovic2020leveraging} introduced BERT for Patents based on the BERT-Large architecture and trained on over 100 million patent abstracts, claims, and descriptions. More recently, researchers have noted that the citation information native to papers and patents can be leveraged to further hone the representations of these transformer-based encoders. The SPECTER/SPECTER2~\citep{cohan2020specter,singh2023scirepeval} models fine-tuned SciBERT through a contrastive triplet loss on true versus negative citations for each paper's title and abstract embedding. SciNCL~\citep{ostendorff2022neighborhood} softened the sampling of this contrastive task by first embedding the citation network and using the cosine similarity of these citation representations to draw positive and negative examples, resulting in further performance improvements. Both of these models have achieved state-of-the-art performance across a number of tasks based on paper embeddings. Similar efforts have been undertaken for patents, including the PaECTER model~\citep{ghosh2024paecter} which performs the same contrastive margin finetuning on the BERT for Patents model as was done for the SPECTER model, using patent citations as input triplets. 

\section{Chronological text sources for pre-training}

The pre-training text data is drawn from three sources covering scientific papers, patents, and high-quality educational web content. All sources, apart from the patent data (which is English only), are filtered to English using the LID.176 fasttext language model~\citep{joulin2016bag, joulin2016fasttext} using confidence threshold 0.6 or greater. 

\subsection{Scientific paper text} 
We use two sources of scientific paper text, both derived from the Semantic Scholar academic database~\citep{kinney2023semantic}. We source full-paper text from a 2026-03-13 snapshot of S2ORC~\citep{lo2020s2orc}, taking only the paper's body text. We refer to this data source as S2ORC. To supplement this full-paper text, we also include the abstracts of all papers in the Semantic Scholar database with reported abstracts, derived from the 2026-02-10 snapshot of the Semantic Scholar Academic Graph. We refer to this source as Abstracts. See Table~\ref{tab:corpus} for information on the token distribution of these sources. The production timestamp for both of these data sources corresponds to the year of publication recorded by Semantic Scholar. 

\subsection{Patent text} 
We use the pre-grant long text of U.S. patent applications provided by PatentsView via the United States Patent and Trademark Office. We pre-train on all pre-grant patent claims, detailed description text, summary text, and figure descriptions available in the pre-grant applications as of the 2026-04-10 snapshot. We refer to this data source as USPTO. The production timestamp for the USPTO data corresponds to the patent's application year. See Table~\ref{tab:corpus} for information on token distribution of the USPTO source. 

\subsection{High-quality web text}
To introduce additional textual diversity, we also pre-train on high-quality educational web text from the FineWeb-Edu corpus~\citep{penedo2024fineweb}, a 1.3 trillion token collection of educational text filtered from FineWeb, a quality-filtered subset of the Common Crawl web archives. We refer to this data source as FineWeb-Edu. The production timestamp for FineWeb-Edu documents corresponds to the Common Crawl snapshot year. These snapshots begin in 2013, so they dictate the first year for which we have reliable record dates across all three corpora. Note that although the FineWeb dataset is deduplicated within each Common Crawl snapshot~\citep{penedo2024fineweb}, the same document may appear multiple times across snapshots. See Table~\ref{tab:corpus} for information on the token distribution of the FineWeb-Edu corpus. 

\section{Chronologically consistent pre-training}

We construct a family of point-in-time encoders for scientific and technological text by thresholding pre-training data according to the latest possible time the underlying text could have been produced. 
For each year $y \in \{2013, 2014, \dots, 2025\}$, we train a ModernBERT model on the corpus $C_{\leq y}$ consisting of all documents with production timestamp no later than $y$. We refer to each of these chronological models as SciTBERT-$y$. For example, SciTBERT-2015 is trained only on paper, patent, and web text with production timestamps in year 2015 or earlier. Thus, the language model SciTBERT-$y$ represents the model a researcher could have trained at the end of year $y$ if immediate snapshots of these sources were available. 

Each SciTBERT-$y$ instantiation is based on the ModernBERT-base architecture~\citep{warner2025smarter}, a 22-layer transformer architecture with 149 million parameters that incorporates local and global alternating attention, rotary position embeddings, and flash attention. Weights are initialized from a zero-mean truncated normal distribution following ModernBERT's depth-scaled scheme~\citep{warner2025smarter}. Input sequences are capped at 512 tokens for more direct comparison with previous scientific encoder models.

We train a byte-pair-encoding tokenizer~\citep{sennrich2016neural} with a vocabulary size of 50,368 on the combined S2ORC, paper abstract, and patent text only. To respect the no-lookahead constraints, a unique tokenizer is trained for each year $y$ and is associated to each model SciTBERT-$y$. For example, the 2018 tokenizer is trained on the paper and patent text in $C_{2018}$, the subword distribution of all paper and patent text in the corpus with production timestamp in or before 2018. Note that these tokenizer differences between SciTBERT vintages make their encoded representations incomparable across years. 

Each SciTBERT-$y$ is trained independently from scratch: for each cutoff $y$ we instantiate a fresh ModernBERT-base model and train its tokenizer based on $C_{\leq y}$. Each model is trained on $C_{\leq y}$ via the traditional BERT masked language modeling task but with a 30\% masking rate~\citep{warner2025smarter}. We use a trapezoidal warmup-stable-decay learning rate schedule~\citep{hu2024minicpm} with 2.1\% warm-up, 85.1\% stable, and 12.8\% cosine decay relative to the token budget for each year (Table~\ref{tab:hparams}). Models are trained using $4{\times}$A100-40GB GPUs with distributed data parallelism in \texttt{bfloat16} precision, a per-device batch of $32$, and gradient accumulation of $8$. We use AdamW \citep{loshchilov2017decoupled} with peak learning rate $3\!\times\!10^{-4}$,
weight decay $0.1$, and gradient clipping at $1.0$. See Table~\ref{tab:hparams} for warmup, stable, and decay token budgets for each SciTBERT model.

\section{Contrastive post-training with chronologically consistent citation data}\label{sec:posttrain}

Citation-informed post-training improves the embeddings of scientific and patent encoders by drawing related documents together in representation space and pushing the embeddings of unrelated documents further apart~\citep{cohan2020specter,ostendorff2022neighborhood,ghosh2024paecter}. Prior methods build their contrastive training examples from a pooled citation graph, so documents published in year $y$ may be paired with documents that cite them after $y$. This leakage of future citation information into the representation of earlier documents can amplify lookahead bias in the latent space. To address this issue, we post-train each SciTBERT-$y$ on a joint citation graph of papers and patents that contains only the citations observable at the end of year $y$. We refer to the resulting model as SciTBERT-$y$-CI. Our chronologically consistent post-training approach broadly follows the neighborhood contrastive approach of SciNCL \citep{ostendorff2022neighborhood}, which we extend to a chronologically consistent citation graph that spans both papers and patents.

\subsection{Point-in-time citation graphs}\label{sec:posttrain-graph}

The graph has three types of directed citation edges. Paper-to-paper citations are from the Semantic Scholar Academic Graph snapshot~\citep{kinney2023semantic}. Patent-to-patent citations come from the USPTO citation records of granted, non-withdrawn U.S. utility patents. Patent-to-paper citations are derived from the Reliance on Science patent-to-paper citation dataset~\citep{marx2020reliance} whose OpenAlex works we link to Semantic Scholar papers through DOI. Papers are dated by their Semantic Scholar publication year. Patents are dated by their grant year. An edge enters the year-$y$ graph only if both of its endpoints are dated no later than $y$. This removes forward-citation lookahead biases: no document in the year-$y$ graph is linked to a document that did not yet exist. We keep only documents that can be embedded, namely papers with both a title and an abstract. Patents are filtered to utility patents. The resulting $2025$ graph contains $28.1$ million papers, $8.1$ million patents, and $642$ million citation edges. Table~\ref{app:posttrain-graph} reports the graph at every yearly cutoff.

\subsection{Neighborhood sampling}\label{sec:posttrain-sampling}

As shown by \citet{ostendorff2022neighborhood}, contrastive learning over citation graphs benefits from continuity in the negative sampling procedure. One way to understand this is by noticing that drawing negative samples from the set of papers that are not cited by another does not guarantee that they are not conceptually similar. Rather than treating cited documents as positives directly, we sample positives and negatives from each document's neighborhood based on a continuous embedding of the graph \citep{ostendorff2022neighborhood}. We train node2vec \citep{grover2016node2vec}
embeddings on the year-$y$ undirected citation graph and retrieve the $4{,}000$ nearest neighbors of each training anchor by cosine similarity. Following \citet{ostendorff2022neighborhood}, the five neighbors ranked $21$st to $25$th are the anchor's positives, and the neighbors ranked $3{,}999$th and $4{,}000$th are its hard negatives. We draw $180{,}000$ paper anchors, $120{,}000$ patent anchors, and a further $50{,}000$ patent anchors that cite at least one paper, which leaves about $349{,}000$ anchors after removing duplicates. Each anchor yields five training examples, one per positive. Each contrastive example pairs the anchor and one positive with both hard negatives and one random negative of the anchor's type drawn from the anchor pool. Papers make up $75\%$ of the anchors sampled during training. Due to the relative rarity of patent-to-paper citations, between $1.3\%$ and $1.7\%$ of positives differ in type from their anchor, so post-training mainly sharpens proximity within papers and within patents (Table~\ref{app:posttrain-graph}). However, as we will see in Section~\ref{sec:results}, this heterogeneous post-training leads to substantial cross-domain performance gains. 

\subsection{Contrastive objective}\label{sec:posttrain-objective}

We embed each document by mean-pooling the final SciTBERT-$y$-CI hidden states over the concatenation of its title and abstract and normalizing to unit length. For an anchor embedding $\mathbf{a}_i$ with paired positive embedding $\mathbf{p}_i$, we minimize the InfoNCE loss
\begin{equation}
\mathcal{L} = -\frac{1}{B}\sum_{i=1}^{B}
\log\frac{\exp(\mathbf{a}_i^{\top}\mathbf{p}_i/\tau)}
{\sum_{\mathbf{c}\in\mathcal{C}_i}\exp(\mathbf{a}_i^{\top}\mathbf{c}/\tau)},
\end{equation}
with temperature $\tau = 0.01$. The candidate set $\mathcal{C}_i$ contains every
positive and every explicit negative in the micro-batch of $16$ examples. Gathered across devices, the positives of other anchors serve as additional negatives. We remove from $\mathcal{C}_i$ any candidate that is the anchor itself or one of its positives other than $\mathbf{p}_i$.

\subsection{Training}\label{sec:posttrain-training}

Each SciTBERT-$y$-CI is initialized from SciTBERT-$y$ and keeps its year-matched tokenizer. We train for $3{,}000$ steps using two GPUs at an effective batch of $64$ examples, or $192{,}000$ examples in total, with a peak learning rate of $10^{-5}$ under a warmup-stable-decay schedule. Table~\ref{app:posttrain-hparams} lists additional training settings.

\section{PatRepEval}\label{sec:patrepeval}

We introduce PatRepEval, a benchmark of 27 tasks across three formats (classification, regression, and proximity) for evaluating representations of patents and the scientific
papers on which they rely. PatRepEval complements SciRepEval~\citep{singh2023scirepeval} with tasks that span patents, papers, and the links between them---including nine tasks that take scientific papers as input and five retrieval tasks whose queries and candidates fall on opposite sides of the paper--patent boundary. Labels are drawn from administrative patent records and established science-of-science datasets, reflecting how patent and paper representations are used to study technological innovation.
Existing benchmarks for scientific document representations, including SciDocs~\citep{cohan2020specter} and SciRepEval~\citep{singh2023scirepeval}, evaluate representations of papers only. To our knowledge, no comparable benchmark evaluates representations of patents alongside papers, or the ability of embedding representations to relate the two. The tasks in PatRepEval are summarized in Table~\ref{tab:patrepeval}, and we describe each task format in brief below. Full details of each task are provided in Appendix~\ref{app:patrepeval}.

All tasks represent a document by the concatenation of its title and abstract. Patent text for all tasks except the Grant Abandon task is drawn from the granted U.S. utility patents published by the USPTO, restricted to patents granted between 2001 and 2025. The tasks built from patent records share a common sample of 1,301,889 granted patents, stratified by grant year and primary Cooperative Patent Classification (CPC) section. This common sample also supplies the candidate documents for patent retrieval. Paper text is drawn from the 2026-03-30 snapshot of OpenAlex \citep{priem2022openalex}, restricted to works linked to U.S. patents by patent-to-paper citations, patent–paper pairs, or Pasteur's quadrant researchers. Each abstract is reconstructed from its inverted index.

\begin{table}[t]
\centering
\small
\begin{tabular}{llrrll}
\toprule
Format & Name & Train & Test & Metric & Source \\
\midrule
\multirow{12}{*}{CLF} & CPC Section & 1{,}041{,}520 & 260{,}369 & Macro F1 & USPTO \\
 & CPC Subclass & 683{,}960 & 170{,}978 & Macro F1 & USPTO \\
 & Assignee Type & 940{,}533 & 235{,}121 & Macro F1 & USPTO \\
 & Academic Public & 182{,}955 & 45{,}670 & Binary F1 & OCPB \\
 & US Foreign & 1{,}777{,}085 & 444{,}143 & Binary F1 & OCPB \\
 & VC Backed & 804{,}792 & 201{,}106 & Binary F1 & OCPB \\
 & Grant Abandon & 1{,}586{,}397 & 396{,}557 & Binary F1 & USPTO \\
 & Renewal & 695{,}027 & 173{,}748 & Binary F1 & USPTO \\
 & PPP Patent & 183{,}218 & 45{,}718 & Binary F1 & PPP \\
 & PPP Paper$^\dagger$ & 233{,}488 & 58{,}252 & Binary F1 & PPP \\
 & PQR Patent & 160{,}141 & 39{,}905 & Binary F1 & PQR \\
 & PQR Paper$^\dagger$ & 160{,}570 & 39{,}658 & Binary F1 & PQR \\
\midrule
\multirow{4}{*}{RGN} & Grant Year & 1{,}041{,}520 & 260{,}369 & Kendall's $\tau$ & USPTO \\
 & Forward Citation & 825{,}530 & 206{,}374 & Kendall's $\tau$ & USPTO \\
 & Kogan Value & 1{,}649{,}143 & 412{,}175 & Kendall's $\tau$ & KPSS \\
 & Science Uptake$^\dagger$ & 800{,}711 & 199{,}393 & Kendall's $\tau$ & RoS \\
\midrule
\multirow{11}{*}{PRX} & Same Inventor & --- & \textbf{Q:} 5{,}304 \textbf{C:} 100{,}000 & MAP, nDCG@10 & USPTO \\
 & Patent Cite & --- & \textbf{Q:} 2{,}810 \textbf{C:} 100{,}000 & MAP, nDCG@10 & USPTO \\
 & Patent Cocite & --- & \textbf{Q:} 2{,}814 \textbf{C:} 100{,}000 & MAP, nDCG@10 & USPTO \\
 & Same Initial Assignee & --- & \textbf{Q:} 1{,}989 \textbf{C:} 100{,}000 & MAP, nDCG@10 & OCPB \\
 & Assignee Patent Match & --- & \textbf{Q:} 500 \textbf{C:} 100{,}000 & MAP & USPTO \\
 & Patent Paper Citation$^\dagger$ & --- & \textbf{Q:} 10{,}926 \textbf{C:} 500{,}000 & MAP, nDCG@10 & RoS \\
 & Paper Patent Retrieval$^\dagger$ & --- & \textbf{Q:} 12{,}342 \textbf{C:} 100{,}000 & MAP, nDCG@10 & RoS \\
 & Papers Cocite$^\dagger$ & --- & \textbf{Q:} 13{,}659 \textbf{C:} 100{,}000 & MAP, nDCG@10 & RoS \\
 & PPP Paper Patent$^\dagger$ & --- & \textbf{Q:} 1{,}117 \textbf{C:} 100{,}000 & MAP, nDCG@10 & PPP \\
 & PQR Patent Paper$^\dagger$ & --- & \textbf{Q:} 13{,}282 \textbf{C:} 500{,}000 & MAP, nDCG@10 & PQR \\
 & PQR Paper Patent$^\dagger$ & --- & \textbf{Q:} 64{,}173 \textbf{C:} 100{,}000 & MAP, nDCG@10 & PQR \\
\bottomrule
\end{tabular}
\caption{Summary of PatRepEval tasks across the three formats: classification (CLF),
regression (RGN), and proximity (PRX). Proximity tasks are evaluated by retrieval over a
corpus, so they have \textbf{Q} queries ranked against a corpus of \textbf{C}
documents and no training split. Tasks marked $^\dagger$ take scientific papers as
input; all other tasks take patents. Label sources are USPTO: U.S.\ Patent and Trademark Office; KPSS: \citet{kogan2017technological}; RoS: Reliance on Science \citep{marx2020reliance}; PPP: patent--paper pairs \citep{marx2020reliance}; PQR: Pasteur's quadrant researchers \citep{scharfmann2025pasteur}; OCPB: assignee attributes \citep{ewens2024firm}.}
\label{tab:patrepeval}
\end{table}

\subsection{Classification}
Assigning patents to technology, ownership, and outcome categories is a basic operation in studies of innovation. PatRepEval includes twelve classification tasks that mimic these operations. Two tasks classify a patent's technology area: its primary CPC section (CPC Section) and, as a multi-label task, its subclasses among the 50 most frequent CPC subclasses (CPC Subclass). Four tasks classify a patent's owner: the type (firm, individual, or government and nonprofit) of its first assignee (Assignee Type), and whether the assignee is an academic or public institution (Academic Public), a U.S. organization (US Foreign), or a venture-backed firm (VC Backed)~\citep{ewens2024firm}. Two tasks classify outcomes: whether a pre-grant application is granted or abandoned (Grant Abandon), and whether a granted patent is maintained through its eighth year according to USPTO maintenance fee records (Renewal). The remaining four tasks capture links between patents and science. We refer to a patent and a paper that disclose the same invention as a patent--paper pair (PPP) \citep{marx2020reliance}, and to a scientist who is also named as an inventor on U.S.\ patents as a Pasteur's quadrant researcher (PQR)~\citep{scharfmann2025pasteur}. These tasks ask whether a patent, or a paper, belongs to a PPP with a match score of at least 3 (PPP Patent, PPP Paper), and whether it was written by a PQR identified with a confidence of at least 0.7 (PQR Patent, PQR Paper). We evaluate embeddings on these classification tasks by training a linear probe on the frozen training embeddings and scoring its test predictions with binary F1 for two-class tasks and macro F1 otherwise.

\subsection{Regression}
The four regression tasks predict quantities associated with the timing, value, and influence of a document: a patent's grant year (Grant Year), the number of distinct patents citing it within five years of grant (Forward Citation), its estimated market value (Kogan Value)~\citep{kogan2017technological}, and,
for a paper, the number of distinct U.S.\ utility patents that cite it (Science Uptake)~\citep{marx2020reliance}. Count and value targets are log-transformed. We evaluate embeddings on regression by fitting a ridge regression, with its regularization chosen on held-out training data, and report Kendall's $\tau$ between the predicted and true values.

\subsection{Proximity}
Proximity tasks rank candidate documents by their relatedness to a query document. This task family mimics operations underlying prior-art search, technology mapping, and relatedness measures between patents and papers. Five tasks relate patents to patents: retrieving the patents that share an inventor with the query (Same Inventor), the patents it cites (Patent Cite), the patents co-cited with it by at least two later patents (Patent Cocite), the patents sharing its initial assignee (Same Initial Assignee), and the patents held by an assignee given the mean embedding of a sample of that assignee's other patents (Assignee Patent Match). The remaining six tasks involve papers. Using the patent-to-paper citations of \citet{marx2020reliance}, we retrieve the papers cited by a patent (Patent Paper Citation), the patents citing a paper (Paper Patent Retrieval), and the papers co-cited with a paper by at least two patents (Papers Cocite). Using the PPP and PQR links, we retrieve the patent paired with a paper (PPP Paper Patent), and the papers or patents produced by the PQR behind a query patent or paper (PQR Patent Paper, PQR Paper Patent). We evaluate embeddings on proximity by ranking every document in a fixed corpus of 100,000 or 500,000 documents by cosine similarity to the query, and report mean average precision (MAP) and nDCG@10 against binary relevance labels.

For the classification and regression tasks we assign 20\% of the examples from each year to the test set, so that the test set preserves the temporal distribution of the full task, and most proximity tasks draw their queries as a year-stratified 1\% sample. We report all PatRepEval scores on a 0--100 scale to match the scaling of SciRepEval. PatRepEval is publicly available (Appendix~\ref{app:patrepeval-sources}). Appendix~\ref{app:patrepeval} also provides further information on the underlying data sources and their licenses.

\section{Evaluation setup}

We evaluate each SciTBERT-$y$ and SciTBERT-$y$-CI model on PatRepEval and SciRepEval. We benchmark these models against a collection of BERT-derived models of similar size. We do not perform any task-specific adaptation or fine-tuning prior to evaluation: the reported performance is derived from each model's mean-pooled embeddings of the task input. 

\subsection{SciRepEval}

SciRepEval~\citep{singh2023scirepeval} is a scientific document embedding benchmark composed of 24 tasks spanning regression, classification, ranking, and search. Models are evaluated by embedding the concatenation of paper title and abstract text for both the train and test split of each task by mean pooling the input representation. Each task evaluation trains a simple model using the training abstract and title embeddings. This task-specific model is then used to predict the test set. We report performance on 18 of SciRepEval's 24 tasks. Citation Prediction is a training-only task, and S2AND requires a separate evaluation pipeline. MeSH Descriptors, SciDocs-MAG, SciDocs-MeSH Diseases, and Paper-Reviewer Matching were omitted for brevity.

\subsection{Baseline models}

We evaluate the performance of a number of similarly sized baseline models commonly employed to embed paper and patent text. SciBERT~\citep{beltagy2019scibert} is a 2019 model based on the BERT-base architecture whose paper-based pre-training data was drawn from the Semantic Scholar database~\citep{kinney2023semantic}. SPECTER2~\citep{singh2023scirepeval} is an extension of SciBERT that was post-trained by contrastive learning over papers within the Semantic Scholar citation network. SPECTER2 is evaluated without any task-specific adapters. SciNCL~\citep{ostendorff2022neighborhood} is also a post-trained extension to SciBERT that uses a continuous representation of the citation network to perform the contrastive sampling. 

ChronoBERT~\citep{he2025chronologically} is a family of ModernBERT-base~\citep{warner2025smarter} architectures that were chronologically trained on a quality-filtered collection of web data. Thus, these models are not necessarily domain-specific to papers or patents. Starting from the first model trained on data from prior to 2000, each successive ChronoBERT model is a training continuation from the prior year's. The ChronoBERT models all use the ModernBERT tokenizer, potentially leaking token distribution information from the future.

PaECTER~\citep{ghosh2024paecter} is post-trained from the BERT for Patents model~\citep{srebrovic2020leveraging}, a BERT-large model pre-trained on global patent data. The PaECTER post-training follows a similar contrastive pattern to the SPECTER2 model, using examiner-added patent citation information to tune embeddings. This model has two to three times the number of parameters of all the other models evaluated in this work. 

\section{Results}\label{sec:results}

The SciTBERT and SciTBERT-CI model families are competitive with, and often outperform, models with similar numbers of parameters that are not chronologically consistent. As depicted in Figure~\ref{fig:per_family_mean_both_benchmarks} and Table~\ref{tab:results_mean_both}, nearly every SciTBERT vintage performs competitively with comparable models across task categories and both benchmarks. The more recent vintages of SciTBERT outperform baselines on regression tasks. SciTBERT outperforms other MLM-only models (SciBERT, ChronoBERT) on mean PatRepEval regression and classification, and trails SciBERT slightly on SciRepEval classification tasks. There are a few vintages that underperform on proximity tasks and one that underperforms in all task categories. However, post-training improves these models’ performance (Section~\ref{sec:contrastive_benefits}).

The SciTBERT and SciTBERT-CI models generally perform competitively, or outperform, across both domains. By contrast, the paper- or patent-specialized baseline models underperform in their non-specialized domains. PaECTER achieves an average of 22.1 on SciRepEval regression tasks, the worst of the baselines and 21\% degradation in performance versus the best SciTBERT/SciTBERT-CI vintages. SciNCL and SPECTER2 achieve mean performance of about 16.0 on PatRepEval proximity tasks compared to 23.9 for the best-performing SciTBERT-CI vintage. These performance differences also reflect the value of PatRepEval and its ability to distinguish model performance across task categories, especially relative to SciRepEval. 

\begin{figure}
    \centering
    \includegraphics[width=\linewidth]{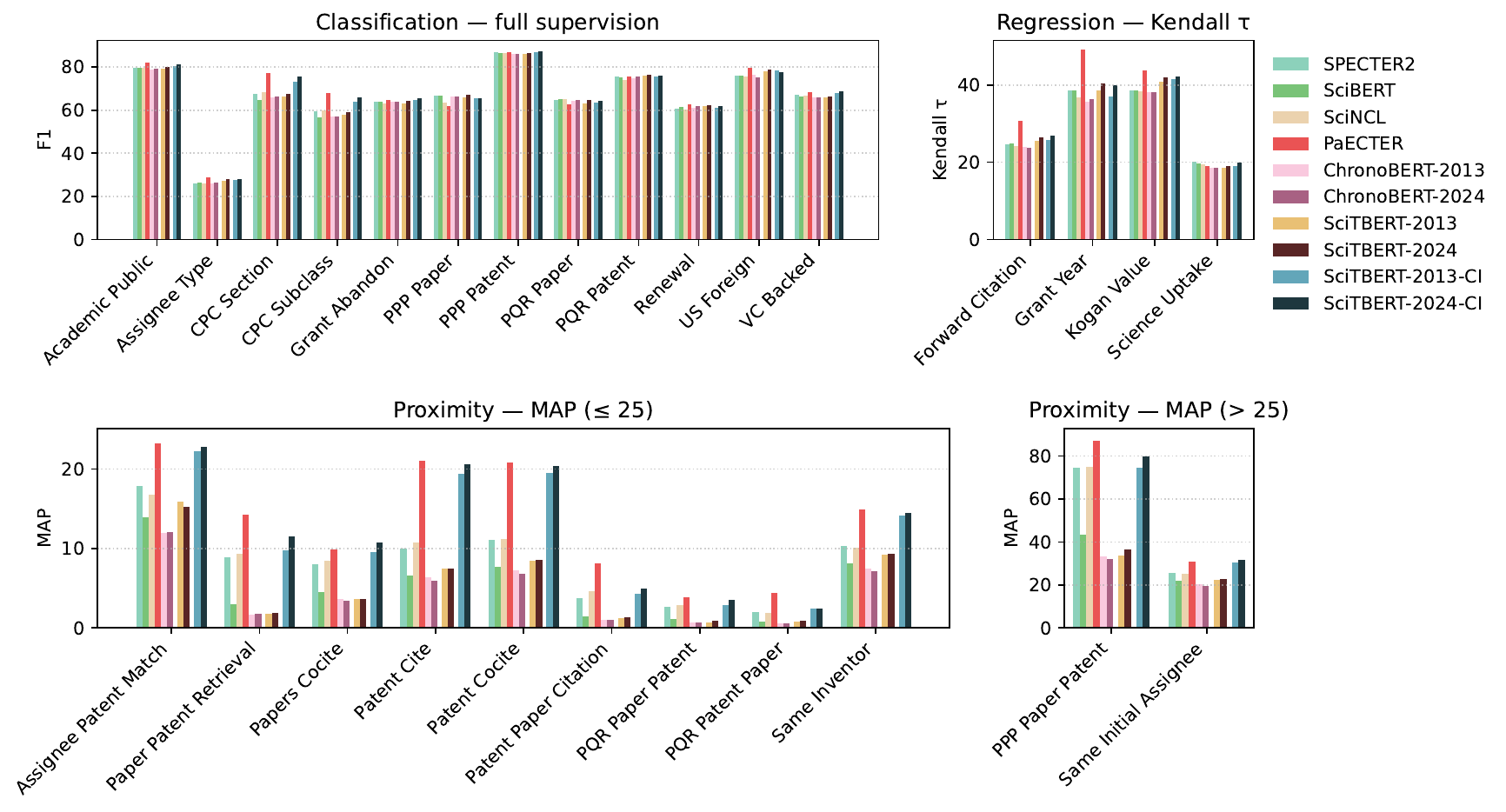}
    \caption{Performance of 2013 and 2024 vintage SciTBERT and SciTBERT-CI models, alongside a number of prior encoder models, across a selection of PatRepEval tasks.}
    \label{fig:combined_patrepeval}
\end{figure}

\begin{figure}
    \centering
    \includegraphics[width=\linewidth]{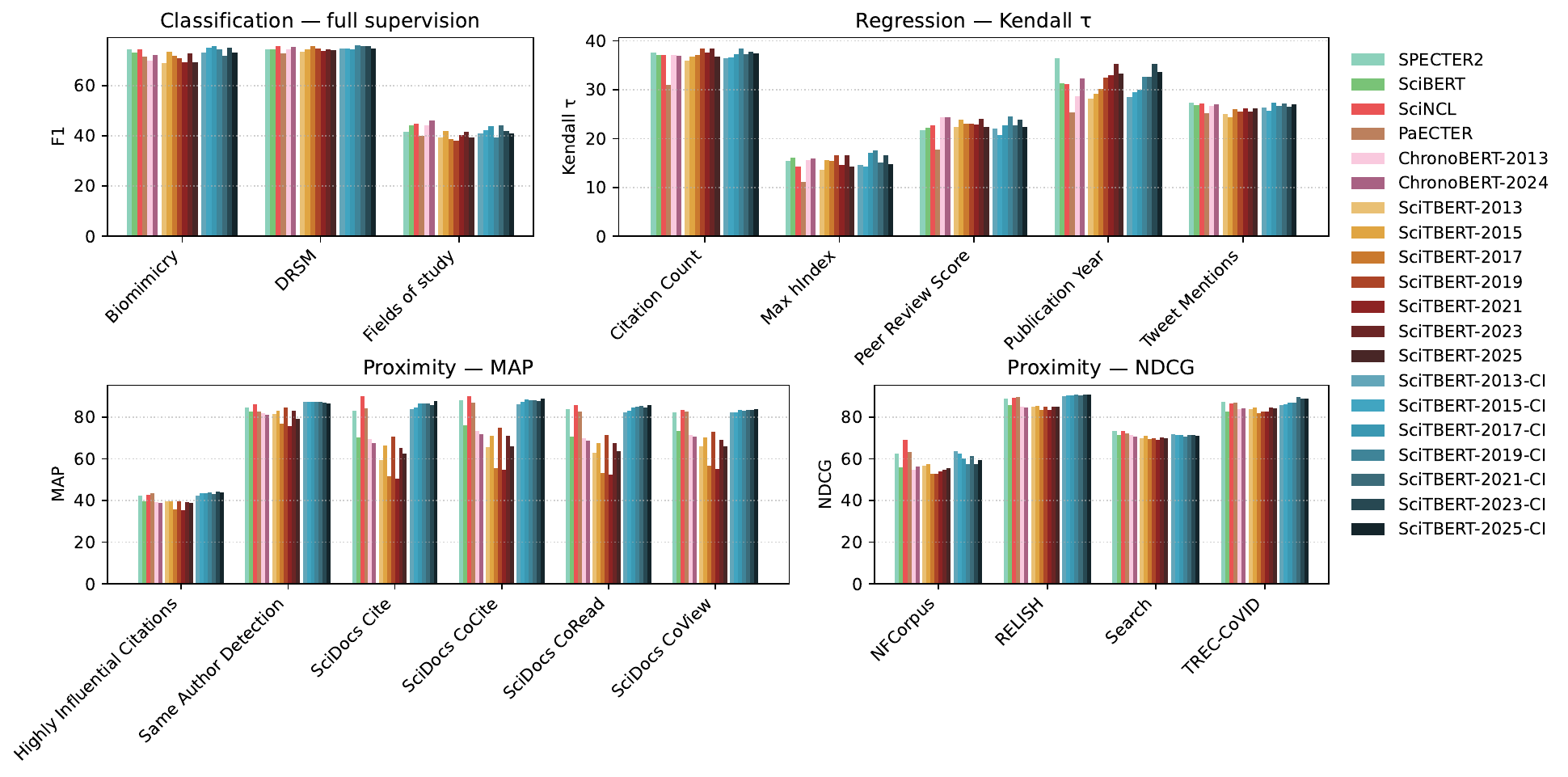}
    \caption{Performance of odd-year SciTBERT and SciTBERT-CI models, alongside a number of baseline models, across a selection of SciRepEval tasks.}
    \label{fig:combined_scirepeval}
\end{figure}

See Figure~\ref{fig:combined_patrepeval} for the performance of the 2013 and 2024 vintages of SciTBERT and SciTBERT-CI compared to baselines on a selection of PatRepEval tasks. The performance of all vintages is given in Tables~\ref{tab:pat_clf} and~\ref{tab:pat_rgn_prx}. The performance for odd-year SciTBERT and SciTBERT-CI vintages alongside baselines on SciRepEval is depicted in Figure~\ref{fig:combined_scirepeval}. The performance of all models can be found in Table~\ref{tab:sci_all}. 

\begin{table*}[t]
\centering
\footnotesize
\setlength{\tabcolsep}{4pt}
\begin{tabular}{lrrrrrrrrr}
\toprule
& \multicolumn{3}{c}{Classification (F1)} & \multicolumn{3}{c}{Regression ($\tau$)} & \multicolumn{3}{c}{Proximity (MAP)} \\
\cmidrule(lr){2-4}\cmidrule(lr){5-7}\cmidrule(lr){8-10}
Model & Sci & Pat & Mean & Sci & Pat & Mean & Sci & Pat & Mean \\
\midrule
SPECTER2 & $63.5$ & $66.1$ & $64.8$ & $\underline{27.7}$ & $30.5$ & $29.1$ & $77.4$ & $15.9$ & $46.7$ \\
SciBERT & $63.9$ & $65.7$ & $64.8$ & $26.7$ & $30.4$ & $28.6$ & $68.8$ & $10.2$ & $39.5$ \\
SciNCL & $\mathbf{65.1}$ & $65.8$ & $65.4$ & $26.5$ & $29.7$ & $28.1$ & $\mathbf{79.7}$ & $16.0$ & $47.8$ \\
PaECTER & $61.4$ & $\mathbf{68.2}$ & $64.8$ & $22.1$ & $\mathbf{35.7}$ & $28.9$ & $77.1$ & $21.7$ & $49.4$ \\
ChronoBERT-2013 & $62.8$ & $65.6$ & $64.2$ & $26.5$ & $29.1$ & $27.8$ & $67.6$ & $8.5$ & $38.1$ \\
ChronoBERT-2024 & $64.6$ & $65.7$ & $65.1$ & $27.3$ & $29.3$ & $28.3$ & $66.5$ & $8.3$ & $37.4$ \\
\midrule
SciTBERT-2013 & $60.5$ & $65.9$ & $63.2$ & $25.0$ & $30.9$ & $28.0$ & $62.5$ & $9.6$ & $36.1$ \\
SciTBERT-2014 & $63.3$ & $66.3$ & $64.8$ & $25.9$ & $31.4$ & $28.7$ & $65.6$ & $9.5$ & $37.5$ \\
SciTBERT-2015 & $63.2$ & $66.6$ & $64.9$ & $26.0$ & $31.6$ & $28.8$ & $66.4$ & $9.8$ & $38.1$ \\
SciTBERT-2016 & $62.7$ & $66.6$ & $64.6$ & $25.7$ & $31.8$ & $28.8$ & $63.6$ & $8.7$ & $36.2$ \\
SciTBERT-2017 & $62.1$ & $66.4$ & $64.3$ & $26.4$ & $31.7$ & $29.0$ & $55.0$ & $6.2$ & $30.6$ \\
SciTBERT-2018 & $61.2$ & $65.8$ & $63.5$ & $25.3$ & $30.5$ & $27.9$ & $65.4$ & $11.0$ & $38.2$ \\
SciTBERT-2019 & $61.2$ & $66.8$ & $64.0$ & $27.2$ & $31.9$ & $29.6$ & $69.0$ & $9.2$ & $39.1$ \\
SciTBERT-2020 & $55.4$ & $64.1$ & $59.8$ & $21.9$ & $27.6$ & $24.8$ & $50.2$ & $4.8$ & $27.5$ \\
SciTBERT-2021 & $61.1$ & $66.5$ & $63.8$ & $26.9$ & $31.8$ & $29.4$ & $54.0$ & $6.0$ & $30.0$ \\
SciTBERT-2022 & $62.2$ & $66.5$ & $64.3$ & $27.5$ & $31.6$ & $29.6$ & $63.1$ & $8.7$ & $35.9$ \\
SciTBERT-2023 & $62.9$ & $67.0$ & $64.9$ & $\mathbf{28.0}$ & $32.3$ & $\underline{30.1}$ & $65.9$ & $8.8$ & $37.3$ \\
SciTBERT-2024 & $60.2$ & $66.8$ & $63.5$ & $27.2$ & $31.9$ & $29.6$ & $65.9$ & $9.9$ & $37.9$ \\
SciTBERT-2025 & $60.9$ & $65.7$ & $63.3$ & $26.6$ & $30.9$ & $28.7$ & $62.7$ & $8.5$ & $35.6$ \\
\midrule
SciTBERT-2013-CI & $62.9$ & $67.4$ & $65.2$ & $25.6$ & $30.9$ & $28.3$ & $77.4$ & $22.2$ & $49.8$ \\
SciTBERT-2014-CI & $64.4$ & $67.6$ & $66.0$ & $25.6$ & $31.5$ & $28.6$ & $77.9$ & $\underline{23.7}$ & $50.8$ \\
SciTBERT-2015-CI & $63.9$ & $67.7$ & $65.8$ & $25.4$ & $31.7$ & $28.5$ & $78.0$ & $23.5$ & $50.8$ \\
SciTBERT-2016-CI & $63.3$ & $67.8$ & $65.5$ & $26.2$ & $31.8$ & $29.0$ & $78.6$ & $23.2$ & $50.9$ \\
SciTBERT-2017-CI & $\underline{64.7}$ & $68.0$ & $\mathbf{66.3}$ & $26.9$ & $31.7$ & $29.3$ & $78.9$ & $23.3$ & $51.1$ \\
SciTBERT-2018-CI & $64.2$ & $67.7$ & $66.0$ & $26.4$ & $31.7$ & $29.0$ & $78.9$ & $\mathbf{23.9}$ & $\underline{51.4}$ \\
SciTBERT-2019-CI & $63.4$ & $\underline{68.1}$ & $65.7$ & $\mathbf{28.0}$ & $32.1$ & $30.0$ & $79.1$ & $23.3$ & $51.2$ \\
SciTBERT-2020-CI & $63.0$ & $67.9$ & $65.4$ & $27.2$ & $31.7$ & $29.4$ & $79.0$ & $23.4$ & $51.2$ \\
SciTBERT-2021-CI & $63.9$ & $\underline{68.1}$ & $66.0$ & $27.0$ & $31.8$ & $29.4$ & $79.0$ & $23.6$ & $51.3$ \\
SciTBERT-2022-CI & $63.8$ & $\underline{68.1}$ & $65.9$ & $27.6$ & $32.0$ & $29.8$ & $\underline{79.5}$ & $23.5$ & $\mathbf{51.5}$ \\
SciTBERT-2023-CI & $64.3$ & $68.0$ & $\underline{66.1}$ & $\mathbf{28.0}$ & $\underline{32.4}$ & $\mathbf{30.2}$ & $78.9$ & $23.2$ & $51.0$ \\
SciTBERT-2024-CI & $64.0$ & $\mathbf{68.2}$ & $\underline{66.1}$ & $\mathbf{28.0}$ & $32.2$ & $\underline{30.1}$ & $79.4$ & $23.5$ & $\mathbf{51.5}$ \\
SciTBERT-2025-CI & $63.0$ & $67.8$ & $65.4$ & $27.0$ & $31.5$ & $29.3$ & $79.4$ & $23.2$ & $51.3$ \\
\bottomrule
\end{tabular}
\caption{Mean performance by task category. Sci and Pat are the means over the SciRepEval and
PatRepEval tasks of each category, and the Mean column averages those means. All scores are on a $0$--$100$ scale. The best performance in each column is bold and the second is underlined.}
\label{tab:results_mean_both}
\end{table*}

\subsection{Contrastive Benefits}\label{sec:contrastive_benefits}
Every SciTBERT-CI vintage outperforms every baseline on the proximity tasks on average (Table~\ref{tab:results_mean_both}). In fact, SciTBERT-2013-CI, trained only on text and citations available by 2013, outperforms every baseline on proximity tasks on average. The model achieves SPECTER2-like performance on SciRepEval and outperforms all the baselines in PatRepEval proximity mean performance. All other SciTBERT-CI vintages achieve even higher performance on average in the proximity tasks. Eight of the 13 SciTBERT-CI vintages (2017 and 2019–2025) exceed the highest baseline regression performance. SciTBERT-2013-CI is the only model vintage that underperforms SciNCL in mean classification performance, and all other vintages match or outperform this best baseline. Overall, the SciTBERT-CI family generally matches the SciRepEval performance of the best paper-specialized models (SPECTER2, SciNCL) and matches PaECTER performance on PatRepEval classification. SciTBERT-CI outperforms PaECTER in proximity tasks on average, despite the fact that PaECTER has more than twice the number of parameters. 

Comparing SciTBERT to SciTBERT-CI models, we find that post-training mainly benefits proximity and classification tasks. SciRepEval proximity performance rises from a cross-vintage average of about 62 for SciTBERT to 79 for SciTBERT-CI. PatRepEval proximity similarly rises from about 9 to 23. The post-training performance benefits to classification are mostly derived from CPC technology classification. Post-training has very little effect on regression task performance, implying most of the signal for regression tasks come from pre-training. 

\begin{figure}
    \centering
    \includegraphics[width=\linewidth]{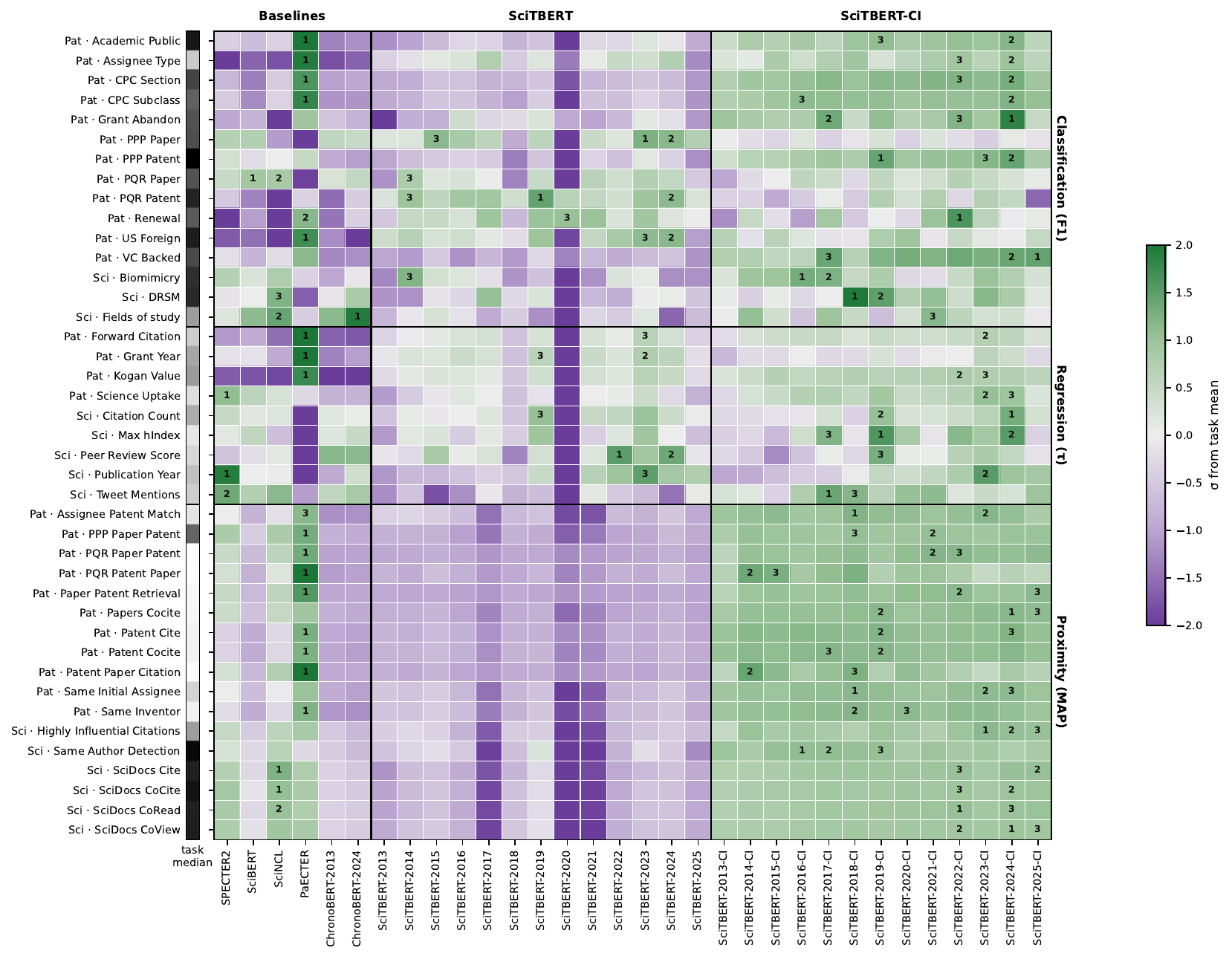}
    \caption{Relative performance of SciTBERT and SciTBERT-CI models versus similar scientific and chronologically timestamped models on all PatRepEval and SciRepEval tasks. Pat indicates PatRepEval tasks and Sci indicates SciRepEval tasks. Performance is reported in standard deviations from the task mean performance across the models. The first-, second-, and third-best performing model for the task is denoted with a 1, 2, or 3, respectively.}
    \label{fig:combined_heatmap_task_zscore}
\end{figure}

There are a number of SciTBERT vintages (2017, 2020, 2021) that show unusually weak performance across particular task categories. The SciTBERT-2020 vintage is a particularly interesting example, lagging mean performance by close to two standard deviations for most tasks (Figure~\ref{fig:combined_heatmap_task_zscore}). This vintage was such an outlier, in fact, that we opted to re-run its entire pre-training process with a new random seed. However, we found a similar lag in performance even with this change of parametrization. We speculate that this underperformance may have more to do with the interaction of the tokenizer and the data distribution at its cutoff year. Regardless, we observe that post-training is able to ameliorate this performance deficit. The SciTBERT-2020-CI model achieves performance across classification, proximity, \textit{and} regression tasks that aligns with its expected performance based on nearby vintages. The fact that contrastive post-training can revive performance across task categories implies that the underperformance in the base model may be related to how the latent space is organized\footnote{We note that, as is the case with BERT models, the base SciTBERT latent space is anisotropic~\citep{ethayarajh2019contextual}. The contrastive post-training over citations reduces this anisotropy.}.

\section{Conclusion}

We introduced SciTBERT, a family of thirteen chronologically consistent ModernBERT-derived models for scientific and technological text, one for each annual cutoff from 2013 to 2025. Each SciTBERT-$y$ model vintage is pre-trained from scratch, with its own tokenizer, on papers, patents, and educational web text produced no later than year $y$. Each comes with a post-trained counterpart, SciTBERT-$y$-CI, tuned on a joint paper and patent citation graph that contains only the citations observable at the end of $y$. To evaluate representations of patents and of the links between patents and science, we also introduced PatRepEval, a benchmark of 27 classification, regression, and proximity tasks, nine of which take papers as input and five of which require explicit retrieval across the paper--patent boundary. Across PatRepEval and SciRepEval, the enforcement of chronological consistency comes at very little performance cost: every SciTBERT-CI vintage exceeds every baseline in mean proximity performance. The SciTBERT-CI family matches the best paper-specialized models on SciRepEval and larger, patent-specialized baseline performance on PatRepEval. While the SciTBERT and SciTBERT-CI models achieve robust cross-domain performance between paper and patent tasks, we find domain-specialized baseline performance to degrade outside the domain the models were trained on. We find citation-informed post-training contributes gains on proximity and technology classification, whereas regression performance is largely driven by pre-training. We release the SciTBERT and SciTBERT-CI models and PatRepEval to support point-in-time measurement in scenarios where the representation of a document should reflect the distribution of knowledge at the time it was produced.


\bibliographystyle{plainnat}
\bibliography{novelty}

\pagebreak

\appendix
\section*{Appendix}
\addcontentsline{toc}{section}{Appendix}
\markboth{Appendix}{Appendix}
\section{Post-training details}\label{app:posttrain}

\subsection{Citation graphs} 
Paper-to-paper edges are the citation records of Semantic Scholar
papers, and patent-to-patent edges are the citations listed on granted, non-withdrawn U.S. utility patents in PatentsView. Patent-to-paper edges are the patent-to-paper citations of the Reliance on Science Patent-to-Paper citations dataset~\citep{marx2020reliance} which identify papers by their OpenAlex identifier. We map each work to its Semantic Scholar identifier by matching on DOI. We drop citations whose work has no matching DOI. Each edge is
stored once, from citing to cited document, and is dated by the later of its two
endpoints. Documents entering the graph at year $y$ remain present in every graph from $y$ onward. Table~\ref{app:posttrain-graph} reports the citation graph properties at each year cutoff.

\subsection{Node2vec and neighbors} 
We train node2vec separately for every cutoff on the undirected version of the year-$y$ graph. Each node2vec model shares the settings in Table~\ref{app:posttrain-hparams}. Nearest neighbors are computed by exact cosine similarity over all node embeddings of the year-$y$ graph, excluding the anchor itself, so that a positive or hard negative of a paper anchor may be a patent and vice versa.

\subsection{Anchors}
The anchor pools are sampled independently. A patent drawn into both patent pools is kept once and counted toward the pool of patents that cite papers. The anchor pools are fixed for each cutoff and are not resampled during training. To reach a paper share of $75\%$, papers are sampled with replacement and patents without replacement within each pass over the anchors. Training stops after $3{,}000$ steps, resulting in each anchor being sampled once in expectation.

\subsection{Negatives} 
The two hard negatives are the anchor's neighbors ranked $3{,}999$th and $4{,}000$th. These negatives are shared for every example with the same anchor. Random (easy) negatives are drawn uniformly from the anchors of the same document type as the anchor, excluding the anchor and its positives. Within the loss, the remaining positives of the same anchor are masked to avoid treatment as negatives.

\begin{table*}[t]
\centering
\small
\begin{tabular}{lrrrrrrrr}
\toprule
& \multicolumn{2}{c}{Nodes (M)} & \multicolumn{3}{c}{Edges (M)} & & \multicolumn{2}{c}{Cross-domain pos.\ (\%)} \\
\cmidrule(lr){2-3}\cmidrule(lr){4-6}\cmidrule(lr){8-9}
Cutoff $y$ & Papers & Patents & S$\to$S & P$\to$P & P$\to$S & Anchors & All & P$\to$S pool \\
\midrule
$2013$ & $2.7$ & $4.6$ & $27.3$ & $49.1$ & $0.6$ & $348{,}242$ & $1.7$ & $3.9$ \\
$2014$ & $3.3$ & $4.9$ & $33.5$ & $54.8$ & $0.7$ & $348{,}316$ & $1.6$ & $3.9$ \\
$2015$ & $3.9$ & $5.1$ & $41.0$ & $60.1$ & $0.8$ & $348{,}534$ & $1.6$ & $3.8$ \\
$2016$ & $4.6$ & $5.4$ & $50.1$ & $65.3$ & $0.9$ & $348{,}557$ & $1.6$ & $4.2$ \\
$2017$ & $5.4$ & $5.7$ & $61.4$ & $71.0$ & $1.0$ & $348{,}568$ & $1.6$ & $4.3$ \\
$2018$ & $6.9$ & $6.0$ & $80.9$ & $76.3$ & $1.1$ & $348{,}662$ & $1.5$ & $4.4$ \\
$2019$ & $9.2$ & $6.3$ & $112.4$ & $83.0$ & $1.3$ & $348{,}790$ & $1.4$ & $4.6$ \\
$2020$ & $12.2$ & $6.6$ & $161.8$ & $89.5$ & $1.5$ & $348{,}863$ & $1.5$ & $4.7$ \\
$2021$ & $15.1$ & $6.9$ & $219.1$ & $95.6$ & $1.7$ & $348{,}862$ & $1.5$ & $4.8$ \\
$2022$ & $18.1$ & $7.2$ & $280.6$ & $101.4$ & $1.9$ & $348{,}956$ & $1.5$ & $4.9$ \\
$2023$ & $21.3$ & $7.5$ & $351.3$ & $107.6$ & $2.2$ & $348{,}943$ & $1.6$ & $5.4$ \\
$2024$ & $24.7$ & $7.8$ & $431.4$ & $113.6$ & $2.2$ & $348{,}991$ & $1.4$ & $4.9$ \\
$2025$ & $28.1$ & $8.1$ & $521.4$ & $118.5$ & $2.2$ & $349{,}001$ & $1.3$ & $4.5$ \\
\bottomrule
\end{tabular}
\caption{The post-training citation graph at each year cutoff $y$. Nodes are papers and patents with a title and an abstract, and an edge is included only if both of its endpoints are dated no later than $y$. The last two columns give the share of positives whose type differs from the anchor's, over all anchors and over anchors drawn from the pool of patents that cite at least one paper.}
\label{app:posttrain-graph}
\end{table*}

\begin{table}[t]
\centering
\small
\begin{tabular}{lr}
\toprule
Setting & Value \\
\midrule
\multicolumn{2}{l}{\textit{node2vec}} \\
\quad Embedding dimension & $128$ \\
\quad Walk length / context size & $20$ / $10$ \\
\quad Walks per node & $10$ \\
\quad Negative samples & $5$ \\
\quad Return / in-out ($p$ / $q$) & $1$ / $1$ \\
\quad Optimizer & Sparse Adam  \\
\quad Learning rate & $0.01$ \\
\quad Epochs & $5$ \\
\quad Batch size & $128$ \\
\midrule
\multicolumn{2}{l}{\textit{Sampling}} \\
\quad Anchor pools (papers / patents / P$\to$S patents) & $180{,}000$ / $120{,}000$ / $50{,}000$ \\
\quad Minimum anchor out-degree & $3$ \\
\quad Paper share of sampled anchors & $0.75$ \\
\quad Positive band (KNN ranks) & $(20, 25]$ \\
\quad Hard-negative band (KNN ranks) & $(3998, 4000]$ \\
\quad Positives per anchor (one example each) & $5$ \\
\quad Hard negatives per example & $2$ \\
\quad Random (easy) negatives per example & $1$ \\ 
\midrule
\multicolumn{2}{l}{\textit{Contrastive training}} \\
\quad Loss & InfoNCE, $\tau = 0.01$ \\
\quad Optimizer & AdamW \\
\quad Learning rate & $10^{-5}$ \\
\quad Weight decay & $0.001$ \\
\quad Gradient clipping & $1$ \\
\quad Schedule (warmup / stable / decay steps) & $500$ / $2000$ / $500$ \\
\quad Steps & $3{,}000$ \\
\quad Effective batch & $64$ \\
\quad Sequence length & $512$ \\
\quad Precision & bfloat16 \\
\bottomrule
\end{tabular}
\caption{Hyperparameter details for the node2vec, sampling, and contrastive learning steps of the SciTBERT-CI post-training. All settings are shared by each SciTBERT-$y$-CI model vintage.}
\label{app:posttrain-hparams}
\end{table}

\begin{table*}[t]
\centering
\small
\begin{tabular}{lrrrrr}
\toprule
Cutoff $y$ & S2ORC & Abstracts & USPTO & FineWeb-Edu & Total \\
\midrule
$2013$ & $10.7$ & $0.8$ & $10.2$ & $23.2$ & $44.9$ \\
$2014$ & $12.9$ & $1.0$ & $10.8$ & $117.3$ & $142.0$ \\
$2015$ & $15.5$ & $1.2$ & $11.4$ & $228.0$ & $256.1$ \\
$2016$ & $18.4$ & $1.5$ & $12.1$ & $329.6$ & $361.6$ \\
$2017$ & $21.8$ & $1.7$ & $12.7$ & $504.2$ & $540.5$ \\
$2018$ & $25.8$ & $2.3$ & $13.4$ & $667.5$ & $708.9$ \\
$2019$ & $30.4$ & $3.1$ & $14.1$ & $827.1$ & $874.6$ \\
$2020$ & $37.3$ & $4.1$ & $14.8$ & $956.7$ & $1012.8$ \\
$2021$ & $45.3$ & $5.1$ & $15.5$ & $1107.9$ & $1173.8$ \\
$2022$ & $52.8$ & $6.0$ & $16.3$ & $1225.2$ & $1300.2$ \\
$2023$ & $60.4$ & $7.1$ & $17.0$ & $1343.6$ & $1428.1$ \\
$2024$ & $67.7$ & $8.2$ & $17.8$ & $1526.7$ & $1620.4$ \\
$2025$ & $74.2$ & $9.4$ & $18.6$ & $1640.0$ & $1742.1$ \\
\bottomrule
\end{tabular}
\caption{Cumulative pretraining corpus $C_{\le y}$ available at each
year cutoff, in billions of tokens. Counts are estimated by scaling each
source's whitespace token count by a byte-pair-encoding factor
calibrated on a $10{,}000$-document sample and by its measured language-ID
retention rate. FineWeb-Edu is reported from $2013$, the first Common Crawl dump ingested, and
paper and patent text from $2000$. Papers and patents predating $2000$ are therefore uncounted in this table but make up a small portion of overall documents.}
\label{tab:corpus}
\end{table*}

\begin{table*}[t]
\centering
\small
\begin{tabular}{lrrrrr}
\toprule
Cutoff $y$ & Warmup & Stable & Decay & Budget & Steps \\
\midrule
$2013$ & $0.5$ & $20.0$ & $3.0$ & $23.5$ & $44{,}821$ \\
$2014$ & $0.6$ & $23.4$ & $3.5$ & $27.5$ & $52{,}450$ \\
$2015$ & $0.7$ & $27.6$ & $4.1$ & $32.4$ & $61{,}797$ \\
$2016$ & $0.7$ & $29.1$ & $4.4$ & $34.2$ & $65{,}230$ \\
$2017$ & $0.8$ & $30.7$ & $4.6$ & $36.1$ & $68{,}853$ \\
$2018$ & $0.8$ & $32.2$ & $4.8$ & $37.8$ & $72{,}096$ \\
$2019$ & $0.8$ & $33.9$ & $5.1$ & $39.8$ & $75{,}911$ \\
$2020$ & $0.9$ & $35.7$ & $5.4$ & $42.0$ & $80{,}107$ \\
$2021$ & $0.9$ & $37.5$ & $5.6$ & $44.0$ & $83{,}922$ \\
$2022$ & $1.0$ & $39.2$ & $5.9$ & $46.1$ & $87{,}928$ \\
$2023$ & $1.0$ & $41.1$ & $6.2$ & $48.3$ & $92{,}124$ \\
$2024$ & $1.1$ & $43.0$ & $6.5$ & $50.6$ & $96{,}510$ \\
$2025$ & $1.1$ & $44.8$ & $6.7$ & $52.6$ & $100{,}326$ \\
\bottomrule
\end{tabular}
\caption{Per-cutoff token budgets, in billions of tokens,
and the resulting optimizer steps. Each budget is set to roughly exhaust the primary scientific and patent sources under the sampling mixture, so it grows with $C_{\le y}$. The phase split is fixed at $2.1\%$ / $85.1\%$ / $12.8\%$ throughout.}
\label{tab:hparams}
\end{table*}

\begin{table*}[t]
\centering
\small
\begin{tabular}{lrrrr}
\toprule
Cutoff $y$ & S2ORC (47.1\%) & Abstracts (11.8\%) & USPTO (35.3\%) & FineWeb-Edu (5.9\%) \\
\midrule
$2013$ & $11.1$ ($1.03\times$) & $2.8$ ($3.36\times$) & $8.3$ ($0.81\times$) & $1.4$ ($0.06\times$) \\
$2014$ & $12.9$ ($1.00\times$) & $3.2$ ($3.23\times$) & $9.7$ ($0.90\times$) & $1.6$ ($0.01\times$) \\
$2015$ & $15.2$ ($0.98\times$) & $3.8$ ($3.15\times$) & $11.4$ ($1.00\times$) & $1.9$ (${<}0.01\times$) \\
$2016$ & $16.1$ ($0.87\times$) & $4.0$ ($2.77\times$) & $12.1$ ($1.00\times$) & $2.0$ (${<}0.01\times$) \\
$2017$ & $17.0$ ($0.78\times$) & $4.2$ ($2.45\times$) & $12.7$ ($1.00\times$) & $2.1$ (${<}0.01\times$) \\
$2018$ & $17.8$ ($0.69\times$) & $4.4$ ($1.96\times$) & $13.3$ ($1.00\times$) & $2.2$ (${<}0.01\times$) \\
$2019$ & $18.7$ ($0.62\times$) & $4.7$ ($1.52\times$) & $14.0$ ($1.00\times$) & $2.3$ (${<}0.01\times$) \\
$2020$ & $19.8$ ($0.53\times$) & $4.9$ ($1.22\times$) & $14.8$ ($1.00\times$) & $2.5$ (${<}0.01\times$) \\
$2021$ & $20.7$ ($0.46\times$) & $5.2$ ($1.02\times$) & $15.5$ ($1.00\times$) & $2.6$ (${<}0.01\times$) \\
$2022$ & $21.7$ ($0.41\times$) & $5.4$ ($0.90\times$) & $16.3$ ($1.00\times$) & $2.7$ (${<}0.01\times$) \\
$2023$ & $22.7$ ($0.38\times$) & $5.7$ ($0.80\times$) & $17.0$ ($1.00\times$) & $2.8$ (${<}0.01\times$) \\
$2024$ & $23.8$ ($0.35\times$) & $6.0$ ($0.72\times$) & $17.9$ ($1.00\times$) & $3.0$ (${<}0.01\times$) \\
$2025$ & $24.8$ ($0.33\times$) & $6.2$ ($0.66\times$) & $18.6$ ($1.00\times$) & $3.1$ (${<}0.01\times$) \\
\bottomrule
\end{tabular}
\caption{Tokens consumed from each source during pretraining, in
billions, with the number of passes over the source in parentheses. The
sampling mixture is fixed, so consumption is the token budget times the source's sampling fraction.}
\label{tab:mixture}
\end{table*}

\section{PatRepEval tasks}\label{app:patrepeval}

Tasks built from patent records alone draw their patents from the common sample described in Appendix~\ref{app:patrepeval-docs} and are dated by grant year (except Grant Abandon which uses filing year), tasks built from external datasets are dated by the patent's filing year, and tasks over papers are dated by paper publication year. Binary tasks whose positive class is rare are balanced by sampling negatives in the same year to match the number of positives in each year. Table~\ref{tab:patrepeval} reports the size of each task.

\subsection{Classification}

\paragraph{CPC Section:} The label is the section (A--H) of the patent's primary CPC
classification. Sections D (textiles) and E (fixed constructions) are smaller than the
others, with 7,001 and 28,142 test examples against roughly 37,500 for each remaining
section.

\paragraph{CPC Subclass:} A multi-label task over the 50 most frequent CPC subclasses in the common patent sample, using all of a patent's CPC assignments. Patents with none of the 50 subclasses are removed.

\paragraph{Assignee Type:} The type of the patent's first-listed assignee, grouped into
firms, individuals, and government and nonprofit organizations. We opt to keep the natural
distribution in which 98.5\% of test patents are assigned to firms.

\paragraph{Academic Public, US Foreign, and VC Backed:} Assignee attributes from
\citet{ewens2024firm} record whether the assignee at filing is a university, hospital, or government body; whether it is a U.S. organization; and whether it is a venture-backed firm, for which positives require a venture-capital match score of at least 7 (of 10). Academic Public and VC Backed use year-matched negatives. US Foreign retains its natural imbalance, with 8,642 non-U.S. against 435,501 U.S. test patents.

\paragraph{Grant Abandon:} The input is the text of a pre-grant application rather than a
granted patent. Positives are applications filed through 2022 that were granted as patents
in the common sample; negatives are utility applications filed in the same years that have
no corresponding grant in the USPTO grant--application crosswalk, matched to the
positives by filing year. Because applications have been published before grant only since
2001, abandoned applications are scarce among the earliest filing years: the classes are
balanced from 2004 onward, while applications filed in or before 2003 (about 11\% of the
test set) are predominantly granted.

\paragraph{Renewal:} Whether a patent was maintained through its eighth year, using the
2026-05-12 release of the USPTO maintenance fee events. A patent is maintained if its year-8 fee was paid or it was reinstated after expiring, and lapsed if it expired for non-payment without reinstatement. Only patents granted in or before 2017 are eligible to ensure the full window is observed. About 35\% of test patents lapsed.

\paragraph{PPP Patent and PPP Paper:} Whether a patent (or paper) appears in a patent--paper pair with a match score of at least 3 on the 1--4 scale of \citet{marx2020reliance}. Negatives are year-matched and drawn from documents that appear in no pair at any score: for patents they are drawn from the common sample, and for papers from the papers cited by U.S. patents \citep{marx2020reliance}.

\paragraph{PQR Patent and PQR Paper:} Whether a patent has an inventor, or a paper has an author, who is matched to a Pasteur's quadrant researcher \citep{scharfmann2025pasteur} with a confidence of at least 0.7. Negatives have no match at any confidence. Both classes are balanced within each year and capped at 100,000 examples per class.

\subsection{Regression}

\paragraph{Grant Year:} The grant year of a patent, drawn from the same patents and split as the CPC Section task.

\paragraph{Forward Citation:} The number of distinct patents citing a patent within five years of its grant. Only patents granted in or before 2020 are included to ensure full window observations. The target is heavy-tailed (median 4, 99th percentile 176) and is transformed as $\log(1+x)$ before fitting.

\paragraph{Kogan Value:} The log of one plus the real market value of a patent from the 2024 release\footnote{Downloaded from \url{https://github.com/KPSS2017/Technological-Innovation-Resource-Allocation-and-Growth-Extended-Data}} of the data of \citet{kogan2017technological}, which covers patents assigned to publicly traded firms. About 60\% of these patents can be joined to granted patent text.

\paragraph{Science Uptake:} The log of one plus the number of distinct U.S. utility patents citing a paper \citep{marx2020reliance}. Every paper is cited at least once by a patent, and we sample 1,000,000 papers in proportion to their publication years.

\subsection{Proximity}

\paragraph{Same Inventor:} Queries are patents with at least three inventors. Relevant documents are the other patents in the common patent sample that share at least one disambiguated inventor with the query.

\paragraph{Patent Cite and Patent Cocite:} Queries are patents citing at least three patents in the common sample. Relevant documents for the Patent Cite task are the cited patents. For the Patent Cocite task the relevant documents are the patents co-cited with the query by at least two distinct citing patents.

\paragraph{Same Initial Assignee:} Relevant documents share the query's initial assignee \citep{ewens2024firm}. To avoid queries dominated by the largest corporate portfolios, we keep assignees with between 3 and 100 patents and sample at most 2,000 queries.

\paragraph{Assignee Patent Match:} We sample 500 assignees with at least eight patents in the common sample and split each assignee's patents 70/30. The query is the mean embedding of the 70\% share, and the relevant documents are the held-out 30\%. We report P@5 and P@10 in addition to MAP.

\paragraph{Patent Paper Citation, Paper Patent Retrieval, and Papers Cocite:} These tasks are all built from the patent-to-paper citations of \citet{marx2020reliance}. Patent Paper Citation retrieves the papers cited by a patent that cites at least three papers. Paper Patent Retrieval retrieves the patents citing a paper that is cited by at least three patents. Papers Cocite retrieves the papers co-cited with a query paper by at least two patents.

\paragraph{PPP Paper Patent:} Queries are papers in a patent--paper pair with a match score of at least 3~\citep{marx2020reliance}, and the relevant documents are the paired patents.

\paragraph{PQR Patent Paper and PQR Paper Patent:} Relevant documents for PQR Patent Paper are the papers written by a Pasteur's quadrant researcher who is an inventor of the
query, using matches with a confidence of at least 0.7~\citep{scharfmann2025pasteur}. Relevant documents for PQR Paper Patent are the patents written by a Pasteur's quadrant researcher who is an author of the query, using the same confidence threshold. 

\subsection{Documents}\label{app:patrepeval-docs}

The common patent sample is drawn from granted, non-withdrawn U.S. utility patents with grant years between 2001 and 2025. We stratify by grant year and primary CPC section, targeting 7,500 patents in each of the $25\times 8$ cells. We drop patents without an abstract, leaving 1,301,889 patents in the shared patent sample. Paper text comes from the OpenAlex works linked to U.S. patents by patent-to-paper citations, patent--paper pairs, or Pasteur's quadrant researchers. Roughly 14 million of these papers have a title, an abstract, and a publication year and enter the tasks.

\subsection{Evaluation details}\label{app:patrepeval-eval}

For classification and regression, we assign 20\% of each year's examples
to the test set and retain the rest for the training set. Years with fewer than 200 examples are kept entirely in the training set. Proximity queries are drawn as a year-stratified 1\% sample except where noted above, and queries are excluded from the corpus. For the Paper Patent Retrieval, Papers Cocite, and both PQR retrieval tasks, each query's negatives are the relevant documents of other queries. Otherwise, the task corpus is filled with randomly sampled patents or papers.

Embeddings are L2-normalized and standardized with the training set mean and standard deviation of each dimension. For classification, the probe is a single
linear layer trained with AdamW for 50 epochs (learning rate $10^{-2}$, weight decay $10^{-4}$, batch size 4,096) using cross-entropy with balanced class weights for single-label tasks and binary cross-entropy for CPC Subclass, and reported scores are the mean over three probe seeds. The classification tasks also define few-shot settings with $k$ examples per class. For simplicity, we do not report these results in this work, but the tasks are available and evaluated by default by the benchmark suite. For regression, the probe is a closed-form ridge regression on min--max scaled targets (log-transformed for Forward Citation), whose penalty is chosen from $\{10^{-4},\dots,10\}\times n$ by Kendall's $\tau$ on a held-out 10\% of the training set (at most 100,000 examples) before refitting on the full training set. Each proximity query is scored against every corpus document by cosine similarity. We compute MAP and nDCG@10 over the 200 highest-ranked documents for each query, using the pytrec\_eval library~\citep{van2018pytrec_eval}.

\subsection{Sources, licenses, and availability}\label{app:patrepeval-sources}

PatRepEval is available on the Hugging Face Hub at \url{https://huggingface.co/datasets/gebhart/patrepeval}, which hosts the train and test
splits of all 27 tasks together with the documents from which they are built. Because several assignee labels derive from data distributed under a non-commercial license (Table~\ref{tab:patrepeval-sources}), PatRepEval is released under the Creative Commons
Attribution-NonCommercial 4.0 International License (CC BY-NC 4.0). Patent records are
drawn from PatentsView and the USPTO and are used under CC BY 4.0 and as U.S.\ government
works, respectively. The code for building the tasks and evaluating models will be released at a later date.

\begin{table}[h]
\centering
\small
\begin{tabular}{p{0.38\linewidth}p{0.42\linewidth}l}
\toprule
Source & Used for & License \\
\midrule
USPTO (via PatentsView) & patent text; CPC, assignee, inventor, citation, and maintenance fee labels & CC BY 4.0 \\
\citet{kogan2017technological} & Kogan Value & None stated \\
Reliance on Science \citep{marx2020reliance} & patent-to-paper citations & CC BY-NC 4.0 \\
Patent--paper pairs \citep{marx2020reliance} & PPP tasks & CC BY-NC 4.0 \\
Pasteur's quadrant researchers \citep{scharfmann2025pasteur} & PQR tasks & CC BY-NC 4.0 \\
Assignee attributes \citep{ewens2024firm} & ownership tasks, Same Initial Assignee & CC BY-NC 4.0 \\
OpenAlex \citep{priem2022openalex} & paper text & CC0 \\
\bottomrule
\end{tabular}
\caption{Data sources used to construct PatRepEval and the terms under which each is
distributed. Most of the USPTO data is derived from PatentsView~\citep{toole2021patentsview}.}
\label{tab:patrepeval-sources}
\end{table}

\begin{table*}[t]
\centering
\footnotesize
\setlength{\tabcolsep}{2pt}
\begin{tabular}{lrrrrrrrrrrrrrrrrrr}
\toprule
& \multicolumn{3}{c}{Classification (F1)} & \multicolumn{5}{c}{Regression (Kendall $\tau$)} & \multicolumn{6}{c}{Proximity (MAP)} & \multicolumn{4}{c}{Proximity (NDCG)} \\
\cmidrule(lr){2-4}\cmidrule(lr){5-9}\cmidrule(lr){10-15}\cmidrule(lr){16-19}
Model & \rotatebox{90}{\strut Biomimicry} & \rotatebox{90}{\strut DRSM} & \rotatebox{90}{\strut Fields of study} & \rotatebox{90}{\strut Citation Count} & \rotatebox{90}{\strut Max hIndex} & \rotatebox{90}{\strut Peer Review Score} & \rotatebox{90}{\strut Publication Year} & \rotatebox{90}{\strut Tweet Mentions} & \rotatebox{90}{\strut Highly Influential Citations} & \rotatebox{90}{\strut Same Author Detection} & \rotatebox{90}{\strut SciDocs Cite} & \rotatebox{90}{\strut SciDocs CoCite} & \rotatebox{90}{\strut SciDocs CoRead} & \rotatebox{90}{\strut SciDocs CoView} & \rotatebox{90}{\strut NFCorpus} & \rotatebox{90}{\strut RELISH} & \rotatebox{90}{\strut Search} & \rotatebox{90}{\strut TREC-CoVID} \\
\midrule
SPECTER2 & $74.3$ & $74.4$ & $41.7$ & $37.6$ & $15.4$ & $21.7$ & $\mathbf{36.4}$ & $\mathbf{27.4}$ & $42.4$ & $84.8$ & $83.2$ & $88.0$ & $83.7$ & $82.4$ & $62.7$ & $88.9$ & $\underline{73.2}$ & $87.3$ \\
SciBERT & $73.0$ & $74.5$ & $44.0$ & $37.1$ & $16.1$ & $22.3$ & $31.3$ & $26.8$ & $39.5$ & $82.7$ & $70.4$ & $76.0$ & $70.8$ & $73.3$ & $56.0$ & $85.6$ & $71.3$ & $82.8$ \\
SciNCL & $74.5$ & $75.8$ & $\underline{44.9}$ & $37.1$ & $14.3$ & $22.8$ & $31.2$ & $27.2$ & $42.7$ & $86.1$ & $\mathbf{90.0}$ & $\mathbf{89.9}$ & $\underline{85.9}$ & $83.6$ & $\mathbf{69.2}$ & $89.1$ & $\mathbf{73.4}$ & $86.6$ \\
PaECTER & $71.4$ & $72.8$ & $40.1$ & $30.9$ & $11.2$ & $17.8$ & $25.4$ & $25.1$ & $43.4$ & $82.7$ & $84.2$ & $87.0$ & $82.8$ & $82.8$ & $63.4$ & $89.8$ & $72.1$ & $86.9$ \\
ChronoBERT-2013 & $69.8$ & $74.4$ & $44.1$ & $37.1$ & $15.5$ & $24.4$ & $28.7$ & $26.6$ & $39.4$ & $82.1$ & $69.3$ & $73.2$ & $70.0$ & $71.6$ & $54.8$ & $85.1$ & $71.3$ & $84.1$ \\
ChronoBERT-2024 & $72.2$ & $75.4$ & $\mathbf{46.2}$ & $36.9$ & $16.0$ & $24.3$ & $32.2$ & $26.9$ & $38.7$ & $81.3$ & $67.7$ & $71.8$ & $68.9$ & $70.6$ & $56.3$ & $84.8$ & $70.7$ & $84.4$ \\
\midrule
SciTBERT-2013 & $69.0$ & $73.3$ & $39.2$ & $35.9$ & $13.6$ & $22.4$ & $28.2$ & $25.0$ & $39.6$ & $81.6$ & $59.5$ & $65.6$ & $62.9$ & $66.2$ & $56.5$ & $85.2$ & $70.0$ & $83.9$ \\
SciTBERT-2014 & $75.7$ & $73.3$ & $41.0$ & $37.0$ & $15.5$ & $22.2$ & $29.3$ & $25.6$ & $40.0$ & $82.5$ & $65.7$ & $69.6$ & $66.0$ & $69.5$ & $58.4$ & $85.2$ & $70.7$ & $83.3$ \\
SciTBERT-2015 & $73.3$ & $74.4$ & $41.9$ & $36.7$ & $15.6$ & $23.9$ & $29.2$ & $24.4$ & $39.7$ & $83.0$ & $66.5$ & $71.1$ & $67.6$ & $70.2$ & $57.3$ & $85.4$ & $71.1$ & $84.6$ \\
SciTBERT-2016 & $73.0$ & $74.2$ & $40.8$ & $36.8$ & $14.6$ & $22.7$ & $29.6$ & $25.0$ & $38.5$ & $82.0$ & $62.8$ & $67.7$ & $63.8$ & $67.2$ & $55.2$ & $84.7$ & $69.7$ & $83.6$ \\
SciTBERT-2017 & $71.9$ & $75.6$ & $38.8$ & $37.2$ & $15.4$ & $23.0$ & $30.1$ & $26.1$ & $35.8$ & $76.7$ & $51.5$ & $55.7$ & $53.2$ & $56.8$ & $53.0$ & $83.6$ & $69.7$ & $81.9$ \\
SciTBERT-2018 & $69.3$ & $74.2$ & $40.0$ & $36.0$ & $14.6$ & $20.6$ & $29.8$ & $25.4$ & $39.5$ & $81.4$ & $64.3$ & $70.5$ & $67.0$ & $70.0$ & $56.5$ & $85.1$ & $70.8$ & $84.7$ \\
SciTBERT-2019 & $70.8$ & $74.8$ & $38.0$ & $\underline{38.5}$ & $16.7$ & $23.1$ & $32.5$ & $25.5$ & $39.6$ & $84.5$ & $70.7$ & $74.8$ & $71.5$ & $73.2$ & $52.7$ & $85.1$ & $69.7$ & $82.5$ \\
SciTBERT-2020 & $62.9$ & $71.1$ & $32.2$ & $31.9$ & $10.9$ & $18.8$ & $24.5$ & $23.6$ & $31.7$ & $75.0$ & $46.0$ & $49.6$ & $48.3$ & $50.8$ & $52.5$ & $82.3$ & $69.6$ & $81.3$ \\
SciTBERT-2021 & $69.2$ & $73.8$ & $40.3$ & $37.6$ & $14.7$ & $22.9$ & $33.1$ & $26.3$ & $35.4$ & $75.7$ & $50.4$ & $54.6$ & $52.4$ & $55.3$ & $54.0$ & $83.5$ & $69.2$ & $82.8$ \\
SciTBERT-2022 & $73.1$ & $73.6$ & $39.9$ & $37.7$ & $15.5$ & $\mathbf{24.9}$ & $33.9$ & $25.7$ & $38.7$ & $80.8$ & $62.1$ & $66.6$ & $63.9$ & $66.7$ & $54.5$ & $84.8$ & $69.7$ & $83.2$ \\
SciTBERT-2023 & $72.7$ & $74.5$ & $41.6$ & $38.4$ & $16.6$ & $24.0$ & $35.2$ & $25.5$ & $39.2$ & $83.0$ & $65.3$ & $71.1$ & $67.5$ & $69.2$ & $54.7$ & $84.9$ & $70.3$ & $84.6$ \\
SciTBERT-2024 & $69.2$ & $74.6$ & $37.0$ & $37.5$ & $15.3$ & $\underline{24.7}$ & $33.7$ & $24.8$ & $39.4$ & $82.0$ & $66.3$ & $70.4$ & $67.4$ & $69.7$ & $52.7$ & $85.0$ & $69.8$ & $84.2$ \\
SciTBERT-2025 & $69.2$ & $74.0$ & $39.4$ & $36.8$ & $14.3$ & $22.4$ & $33.3$ & $26.2$ & $38.9$ & $79.3$ & $62.5$ & $66.0$ & $63.5$ & $66.2$ & $55.7$ & $84.9$ & $69.8$ & $84.2$ \\
\midrule
SciTBERT-2013-CI & $73.1$ & $74.6$ & $41.1$ & $36.4$ & $14.6$ & $22.1$ & $28.6$ & $26.4$ & $42.4$ & $87.2$ & $84.0$ & $86.1$ & $82.3$ & $82.3$ & $\underline{63.8}$ & $90.0$ & $71.9$ & $85.8$ \\
SciTBERT-2014-CI & $75.0$ & $74.1$ & $43.9$ & $36.5$ & $14.8$ & $21.8$ & $28.8$ & $26.3$ & $43.8$ & $86.6$ & $84.1$ & $86.8$ & $83.4$ & $82.3$ & $63.1$ & $90.5$ & $71.4$ & $84.6$ \\
SciTBERT-2015-CI & $75.1$ & $74.6$ & $42.1$ & $36.6$ & $14.2$ & $20.7$ & $29.5$ & $25.8$ & $43.3$ & $\underline{87.4}$ & $84.7$ & $87.3$ & $83.1$ & $82.4$ & $62.5$ & $90.5$ & $71.4$ & $86.1$ \\
SciTBERT-2016-CI & $\mathbf{75.9}$ & $74.2$ & $39.6$ & $36.7$ & $15.8$ & $21.7$ & $30.0$ & $26.9$ & $43.6$ & $\mathbf{87.5}$ & $86.1$ & $87.8$ & $83.8$ & $82.9$ & $62.1$ & $90.6$ & $71.6$ & $87.6$ \\
SciTBERT-2017-CI & $\underline{75.8}$ & $74.6$ & $43.7$ & $37.3$ & $17.0$ & $22.6$ & $30.0$ & $\mathbf{27.4}$ & $43.4$ & $\mathbf{87.5}$ & $86.4$ & $88.4$ & $84.5$ & $83.4$ & $60.4$ & $90.5$ & $71.4$ & $87.1$ \\
SciTBERT-2018-CI & $73.8$ & $\mathbf{76.5}$ & $42.4$ & $36.3$ & $15.1$ & $22.1$ & $31.1$ & $\underline{27.3}$ & $43.1$ & $\underline{87.4}$ & $86.6$ & $88.2$ & $84.7$ & $83.5$ & $61.2$ & $90.6$ & $71.5$ & $86.9$ \\
SciTBERT-2019-CI & $74.6$ & $\underline{76.0}$ & $39.5$ & $\underline{38.5}$ & $\mathbf{17.6}$ & $24.6$ & $32.6$ & $26.8$ & $43.8$ & $\mathbf{87.5}$ & $86.7$ & $88.3$ & $85.0$ & $83.2$ & $57.4$ & $90.7$ & $70.8$ & $87.1$ \\
SciTBERT-2020-CI & $71.7$ & $75.3$ & $42.0$ & $37.3$ & $16.5$ & $22.7$ & $32.3$ & $27.1$ & $43.5$ & $86.7$ & $86.5$ & $88.6$ & $85.2$ & $83.6$ & $60.3$ & $90.6$ & $71.0$ & $89.3$ \\
SciTBERT-2021-CI & $71.8$ & $75.6$ & $44.2$ & $37.3$ & $15.1$ & $22.6$ & $32.7$ & $27.2$ & $43.1$ & $87.3$ & $86.6$ & $88.2$ & $85.4$ & $83.5$ & $61.4$ & $90.6$ & $71.4$ & $\underline{89.5}$ \\
SciTBERT-2022-CI & $73.7$ & $75.0$ & $42.7$ & $37.5$ & $16.9$ & $23.6$ & $33.4$ & $26.3$ & $43.8$ & $86.9$ & $\underline{87.6}$ & $88.9$ & $\mathbf{86.1}$ & $\underline{83.9}$ & $61.1$ & $90.7$ & $71.0$ & $88.6$ \\
SciTBERT-2023-CI & $75.2$ & $75.7$ & $42.0$ & $37.9$ & $16.6$ & $23.8$ & $\underline{35.3}$ & $26.6$ & $\mathbf{44.2}$ & $87.0$ & $85.8$ & $87.9$ & $84.7$ & $83.5$ & $57.4$ & $\underline{90.8}$ & $71.5$ & $89.0$ \\
SciTBERT-2024-CI & $74.4$ & $75.4$ & $42.2$ & $\mathbf{38.9}$ & $\underline{17.5}$ & $23.4$ & $33.9$ & $26.4$ & $\underline{44.0}$ & $86.5$ & $87.4$ & $\underline{89.0}$ & $85.7$ & $\mathbf{84.0}$ & $56.0$ & $\underline{90.8}$ & $70.6$ & $\mathbf{89.8}$ \\
SciTBERT-2025-CI & $73.3$ & $74.7$ & $41.0$ & $37.4$ & $14.7$ & $22.3$ & $33.6$ & $27.1$ & $43.9$ & $86.8$ & $\underline{87.6}$ & $88.8$ & $85.6$ & $83.7$ & $59.4$ & $\mathbf{90.9}$ & $71.2$ & $88.8$ \\
\bottomrule
\end{tabular}
\caption{SciRepEval performance, per task. All scores are on a $0$--$100$ scale. Bold denotes best-performing model for the task and underline the second (columnwise).}
\label{tab:sci_all}
\end{table*}

\begin{table*}[t]
\centering
\footnotesize
\setlength{\tabcolsep}{2pt}
\begin{tabular}{lrrrrrrrrrrrr}
\toprule
Model & \rotatebox{90}{\strut Academic Public} & \rotatebox{90}{\strut Assignee Type} & \rotatebox{90}{\strut CPC Section} & \rotatebox{90}{\strut CPC Subclass} & \rotatebox{90}{\strut Grant Abandon} & \rotatebox{90}{\strut PPP Paper} & \rotatebox{90}{\strut PPP Patent} & \rotatebox{90}{\strut PQR Paper} & \rotatebox{90}{\strut PQR Patent} & \rotatebox{90}{\strut Renewal} & \rotatebox{90}{\strut US Foreign} & \rotatebox{90}{\strut VC Backed} \\
\midrule
SPECTER2 & $79.8$ & $25.8$ & $67.3$ & $59.5$ & $63.8$ & $66.6$ & $86.8$ & $64.6$ & $75.5$ & $60.5$ & $75.9$ & $66.9$ \\
SciBERT & $79.6$ & $26.3$ & $64.7$ & $56.7$ & $63.9$ & $66.6$ & $86.4$ & $\mathbf{65.0}$ & $75.1$ & $61.4$ & $76.1$ & $66.3$ \\
SciNCL & $79.8$ & $26.2$ & $68.4$ & $60.0$ & $63.0$ & $63.7$ & $86.6$ & $\underline{64.9}$ & $73.9$ & $60.3$ & $75.6$ & $66.9$ \\
PaECTER & $\mathbf{81.9}$ & $\mathbf{28.9}$ & $\mathbf{77.1}$ & $\mathbf{68.1}$ & $64.9$ & $61.9$ & $86.8$ & $62.7$ & $75.6$ & $\underline{62.9}$ & $\mathbf{79.5}$ & $68.3$ \\
ChronoBERT-2013 & $79.0$ & $26.2$ & $66.1$ & $57.0$ & $64.0$ & $66.3$ & $86.0$ & $64.5$ & $75.0$ & $61.1$ & $76.4$ & $65.8$ \\
ChronoBERT-2024 & $79.1$ & $26.3$ & $66.4$ & $57.0$ & $63.9$ & $66.2$ & $85.9$ & $64.7$ & $75.5$ & $61.8$ & $75.3$ & $65.9$ \\
\midrule
SciTBERT-2013 & $79.1$ & $27.2$ & $66.5$ & $57.8$ & $63.0$ & $65.8$ & $86.0$ & $63.3$ & $75.9$ & $61.7$ & $78.1$ & $66.1$ \\
SciTBERT-2014 & $79.3$ & $27.4$ & $66.7$ & $58.3$ & $63.8$ & $65.8$ & $86.2$ & $\underline{64.9}$ & $\underline{76.3}$ & $62.4$ & $78.4$ & $66.0$ \\
SciTBERT-2015 & $79.5$ & $27.6$ & $67.9$ & $59.3$ & $63.8$ & $\underline{67.3}$ & $86.2$ & $64.4$ & $76.0$ & $62.4$ & $78.0$ & $66.6$ \\
SciTBERT-2016 & $79.8$ & $27.7$ & $67.9$ & $59.2$ & $64.6$ & $66.8$ & $86.3$ & $64.5$ & $76.2$ & $62.3$ & $78.1$ & $65.9$ \\
SciTBERT-2017 & $79.8$ & $28.0$ & $67.1$ & $58.3$ & $64.1$ & $66.4$ & $86.3$ & $64.2$ & $76.2$ & $62.7$ & $77.8$ & $66.3$ \\
SciTBERT-2018 & $79.4$ & $27.1$ & $67.2$ & $57.5$ & $64.2$ & $63.9$ & $85.7$ & $63.2$ & $75.9$ & $61.6$ & $77.4$ & $66.0$ \\
SciTBERT-2019 & $79.6$ & $27.6$ & $68.1$ & $59.7$ & $64.5$ & $66.4$ & $86.2$ & $64.6$ & $\mathbf{76.5}$ & $62.8$ & $78.7$ & $66.8$ \\
SciTBERT-2020 & $77.8$ & $26.5$ & $63.1$ & $53.8$ & $63.8$ & $58.9$ & $84.6$ & $60.8$ & $76.1$ & $62.8$ & $75.7$ & $65.7$ \\
SciTBERT-2021 & $79.8$ & $27.6$ & $67.3$ & $59.0$ & $63.8$ & $66.2$ & $86.3$ & $64.8$ & $76.0$ & $62.8$ & $78.2$ & $66.4$ \\
SciTBERT-2022 & $79.9$ & $27.8$ & $67.6$ & $58.9$ & $63.9$ & $65.8$ & $86.2$ & $64.5$ & $75.9$ & $62.3$ & $78.6$ & $66.2$ \\
SciTBERT-2023 & $80.2$ & $27.7$ & $68.1$ & $60.0$ & $64.4$ & $\mathbf{67.5}$ & $86.6$ & $64.8$ & $76.2$ & $62.7$ & $78.8$ & $66.4$ \\
SciTBERT-2024 & $80.0$ & $28.0$ & $67.5$ & $59.2$ & $64.2$ & $\underline{67.3}$ & $86.3$ & $64.7$ & $\underline{76.3}$ & $62.2$ & $\underline{78.9}$ & $66.5$ \\
SciTBERT-2025 & $79.4$ & $26.6$ & $65.7$ & $57.0$ & $63.7$ & $66.6$ & $85.8$ & $63.8$ & $75.9$ & $62.1$ & $76.6$ & $65.9$ \\
\midrule
SciTBERT-2013-CI & $80.4$ & $27.7$ & $73.4$ & $64.1$ & $64.9$ & $65.4$ & $86.8$ & $63.5$ & $75.5$ & $61.3$ & $78.4$ & $67.9$ \\
SciTBERT-2014-CI & $80.8$ & $27.6$ & $74.3$ & $64.4$ & $64.8$ & $65.1$ & $86.9$ & $64.0$ & $75.6$ & $62.4$ & $77.5$ & $67.9$ \\
SciTBERT-2015-CI & $80.7$ & $28.1$ & $74.2$ & $64.9$ & $64.8$ & $64.9$ & $87.0$ & $64.2$ & $75.3$ & $62.0$ & $78.4$ & $67.7$ \\
SciTBERT-2016-CI & $80.8$ & $27.8$ & $74.7$ & $65.5$ & $64.8$ & $65.7$ & $87.0$ & $64.7$ & $75.5$ & $61.4$ & $77.9$ & $67.8$ \\
SciTBERT-2017-CI & $80.6$ & $28.0$ & $75.2$ & $65.4$ & $\underline{65.1}$ & $64.8$ & $87.1$ & $64.6$ & $75.8$ & $62.7$ & $77.5$ & $\underline{68.6}$ \\
SciTBERT-2018-CI & $80.9$ & $\underline{28.2}$ & $74.6$ & $65.3$ & $64.6$ & $65.2$ & $87.0$ & $64.0$ & $75.5$ & $61.8$ & $77.5$ & $67.9$ \\
SciTBERT-2019-CI & $81.0$ & $27.7$ & $75.2$ & $65.4$ & $65.0$ & $65.9$ & $\mathbf{87.4}$ & $64.7$ & $76.1$ & $62.1$ & $78.5$ & $68.4$ \\
SciTBERT-2020-CI & $80.8$ & $27.9$ & $75.0$ & $65.0$ & $64.8$ & $64.8$ & $87.1$ & $64.4$ & $75.9$ & $61.9$ & $78.7$ & $68.5$ \\
SciTBERT-2021-CI & $80.9$ & $28.1$ & $75.4$ & $65.2$ & $64.8$ & $65.8$ & $\underline{87.2}$ & $64.6$ & $76.1$ & $62.8$ & $77.6$ & $68.4$ \\
SciTBERT-2022-CI & $81.0$ & $\underline{28.2}$ & $75.4$ & $65.3$ & $65.0$ & $65.1$ & $\underline{87.2}$ & $64.8$ & $75.6$ & $\mathbf{63.2}$ & $78.2$ & $\underline{68.6}$ \\
SciTBERT-2023-CI & $80.9$ & $28.0$ & $75.0$ & $65.3$ & $64.9$ & $64.7$ & $\underline{87.2}$ & $64.6$ & $76.1$ & $62.5$ & $77.8$ & $68.5$ \\
SciTBERT-2024-CI & $\underline{81.1}$ & $\underline{28.2}$ & $\underline{75.8}$ & $\underline{65.8}$ & $\mathbf{65.4}$ & $65.6$ & $\mathbf{87.4}$ & $64.5$ & $76.0$ & $62.1$ & $77.6$ & $\underline{68.6}$ \\
SciTBERT-2025-CI & $80.6$ & $27.9$ & $74.4$ & $65.3$ & $64.6$ & $65.2$ & $87.0$ & $64.3$ & $74.9$ & $62.2$ & $78.2$ & $\mathbf{68.7}$ \\
\bottomrule
\end{tabular}
\caption{PatRepEval classification performance, per task. All scores are on a $0$--$100$ scale. Bold denotes best-performing model for the task and underline the second (columnwise).}
\label{tab:pat_clf}
\end{table*}

\begin{table*}[t]
\centering
\footnotesize
\setlength{\tabcolsep}{2pt}
\begin{tabular}{lrrrrrrrrrrrrrrr}
\toprule
& \multicolumn{4}{c}{Regression ($\tau$)} & \multicolumn{11}{c}{Proximity (MAP)} \\
\cmidrule(lr){2-5}\cmidrule(lr){6-16}
Model & \rotatebox{90}{\strut Forward Citation} & \rotatebox{90}{\strut Grant Year} & \rotatebox{90}{\strut Kogan Value} & \rotatebox{90}{\strut Science Uptake} & \rotatebox{90}{\strut Assignee Patent Match} & \rotatebox{90}{\strut Paper Patent Retrieval} & \rotatebox{90}{\strut Papers Cocite} & \rotatebox{90}{\strut Patent Cite} & \rotatebox{90}{\strut Patent Cocite} & \rotatebox{90}{\strut Patent Paper Citation} & \rotatebox{90}{\strut PPP Paper Patent} & \rotatebox{90}{\strut PQR Paper Patent} & \rotatebox{90}{\strut PQR Patent Paper} & \rotatebox{90}{\strut Same Initial Assignee} & \rotatebox{90}{\strut Same Inventor} \\
\midrule
SPECTER2 & $24.6$ & $38.5$ & $38.7$ & $\mathbf{20.1}$ & $17.8$ & $8.9$ & $8.1$ & $10.0$ & $11.0$ & $3.8$ & $74.6$ & $2.7$ & $2.0$ & $25.5$ & $10.4$ \\
SciBERT & $24.9$ & $38.5$ & $38.6$ & $19.7$ & $13.9$ & $2.9$ & $4.5$ & $6.6$ & $7.7$ & $1.4$ & $43.2$ & $1.1$ & $0.8$ & $21.8$ & $8.1$ \\
SciNCL & $24.1$ & $36.7$ & $38.4$ & $19.5$ & $16.8$ & $9.3$ & $8.4$ & $10.7$ & $11.2$ & $4.6$ & $75.0$ & $2.9$ & $1.9$ & $25.1$ & $10.1$ \\
PaECTER & $\mathbf{30.8}$ & $\mathbf{49.1}$ & $\mathbf{43.8}$ & $19.0$ & $\underline{23.3}$ & $\mathbf{14.3}$ & $9.9$ & $\mathbf{21.0}$ & $\mathbf{20.8}$ & $\mathbf{8.1}$ & $\mathbf{87.3}$ & $\mathbf{3.9}$ & $\mathbf{4.4}$ & $30.9$ & $\mathbf{15.0}$ \\
ChronoBERT-2013 & $23.9$ & $35.7$ & $38.2$ & $18.7$ & $11.9$ & $1.7$ & $3.7$ & $6.3$ & $7.3$ & $1.0$ & $33.1$ & $0.6$ & $0.5$ & $20.5$ & $7.4$ \\
ChronoBERT-2024 & $23.8$ & $36.4$ & $38.1$ & $18.7$ & $12.0$ & $1.7$ & $3.4$ & $5.9$ & $6.8$ & $1.0$ & $32.2$ & $0.7$ & $0.5$ & $19.7$ & $7.1$ \\
\midrule
SciTBERT-2013 & $25.6$ & $38.6$ & $40.8$ & $18.5$ & $15.9$ & $1.8$ & $3.6$ & $7.5$ & $8.4$ & $1.2$ & $33.7$ & $0.7$ & $0.8$ & $22.4$ & $9.2$ \\
SciTBERT-2014 & $26.0$ & $39.4$ & $41.4$ & $18.9$ & $16.1$ & $1.7$ & $3.6$ & $7.0$ & $8.3$ & $1.0$ & $33.8$ & $0.8$ & $0.7$ & $22.1$ & $9.1$ \\
SciTBERT-2015 & $26.3$ & $39.4$ & $41.6$ & $19.2$ & $15.6$ & $1.9$ & $3.9$ & $8.0$ & $8.9$ & $1.4$ & $34.1$ & $0.8$ & $1.0$ & $22.8$ & $9.6$ \\
SciTBERT-2016 & $26.4$ & $39.9$ & $41.5$ & $19.4$ & $14.6$ & $1.5$ & $3.2$ & $7.0$ & $8.0$ & $0.9$ & $28.5$ & $0.7$ & $0.7$ & $21.3$ & $8.9$ \\
SciTBERT-2017 & $26.5$ & $39.6$ & $41.4$ & $19.3$ & $10.6$ & $0.9$ & $2.1$ & $4.8$ & $5.8$ & $0.6$ & $18.4$ & $0.5$ & $0.5$ & $17.2$ & $6.6$ \\
SciTBERT-2018 & $25.3$ & $37.4$ & $40.4$ & $18.8$ & $14.3$ & $1.8$ & $3.5$ & $7.1$ & $8.3$ & $1.1$ & $33.6$ & $0.7$ & $0.6$ & $21.2$ & $8.5$ \\
SciTBERT-2019 & $26.5$ & $40.4$ & $41.8$ & $19.1$ & $14.9$ & $1.6$ & $3.6$ & $7.0$ & $8.4$ & $1.1$ & $31.8$ & $0.7$ & $0.8$ & $22.0$ & $9.3$ \\
SciTBERT-2020 & $23.0$ & $33.2$ & $38.2$ & $15.9$ & $8.9$ & $0.7$ & $1.1$ & $4.3$ & $5.0$ & $0.4$ & $12.8$ & $0.3$ & $0.2$ & $13.5$ & $5.1$ \\
SciTBERT-2021 & $26.5$ & $39.9$ & $41.7$ & $19.3$ & $9.3$ & $1.0$ & $2.1$ & $4.7$ & $5.5$ & $0.7$ & $19.0$ & $0.5$ & $0.5$ & $16.2$ & $6.1$ \\
SciTBERT-2022 & $26.2$ & $39.5$ & $41.5$ & $19.3$ & $14.4$ & $1.4$ & $3.1$ & $7.0$ & $7.8$ & $1.0$ & $29.7$ & $0.7$ & $0.8$ & $21.0$ & $8.7$ \\
SciTBERT-2023 & $\underline{27.0}$ & $\underline{40.7}$ & $42.2$ & $19.6$ & $13.8$ & $1.4$ & $3.4$ & $6.6$ & $7.7$ & $0.9$ & $30.8$ & $0.7$ & $0.7$ & $21.5$ & $8.9$ \\
SciTBERT-2024 & $26.5$ & $40.3$ & $41.9$ & $19.1$ & $15.2$ & $1.9$ & $3.6$ & $7.5$ & $8.6$ & $1.3$ & $36.5$ & $0.9$ & $0.9$ & $22.5$ & $9.3$ \\
SciTBERT-2025 & $25.6$ & $38.4$ & $40.8$ & $18.7$ & $13.2$ & $1.5$ & $3.1$ & $6.3$ & $7.6$ & $0.9$ & $31.4$ & $0.6$ & $0.6$ & $20.6$ & $8.1$ \\
\midrule
SciTBERT-2013-CI & $25.8$ & $37.1$ & $41.6$ & $19.0$ & $22.3$ & $9.7$ & $9.6$ & $19.4$ & $19.5$ & $4.3$ & $74.4$ & $2.9$ & $2.4$ & $30.5$ & $14.1$ \\
SciTBERT-2014-CI & $26.5$ & $38.3$ & $41.8$ & $19.5$ & $22.8$ & $11.5$ & $10.3$ & $20.3$ & $20.2$ & $\underline{6.1}$ & $80.7$ & $3.5$ & $\underline{3.1}$ & $31.1$ & $14.4$ \\
SciTBERT-2015-CI & $26.6$ & $38.3$ & $42.2$ & $19.6$ & $23.1$ & $11.6$ & $10.3$ & $20.0$ & $19.9$ & $5.5$ & $79.3$ & $3.4$ & $3.0$ & $30.8$ & $14.5$ \\
SciTBERT-2016-CI & $26.7$ & $38.9$ & $42.0$ & $19.8$ & $22.2$ & $11.2$ & $10.2$ & $20.3$ & $20.2$ & $4.8$ & $78.4$ & $3.4$ & $2.6$ & $31.2$ & $14.6$ \\
SciTBERT-2017-CI & $26.8$ & $38.2$ & $42.2$ & $19.8$ & $22.5$ & $11.5$ & $10.2$ & $20.3$ & $20.4$ & $5.3$ & $77.5$ & $3.5$ & $2.8$ & $31.2$ & $14.6$ \\
SciTBERT-2018-CI & $26.5$ & $38.3$ & $42.3$ & $19.7$ & $\mathbf{23.5}$ & $11.8$ & $10.4$ & $20.0$ & $20.1$ & $5.7$ & $80.8$ & $3.6$ & $3.0$ & $\mathbf{31.7}$ & $\underline{14.9}$ \\
SciTBERT-2019-CI & $26.7$ & $39.5$ & $42.2$ & $19.9$ & $23.0$ & $11.4$ & $\underline{10.6}$ & $\underline{20.6}$ & $\underline{20.5}$ & $4.7$ & $78.2$ & $3.3$ & $2.4$ & $31.3$ & $14.5$ \\
SciTBERT-2020-CI & $26.6$ & $38.4$ & $42.2$ & $19.6$ & $22.9$ & $11.9$ & $10.0$ & $19.8$ & $19.8$ & $5.4$ & $79.5$ & $3.6$ & $2.7$ & $30.9$ & $14.7$ \\
SciTBERT-2021-CI & $26.6$ & $38.8$ & $42.3$ & $19.7$ & $22.4$ & $11.9$ & $10.2$ & $19.6$ & $19.6$ & $5.1$ & $\underline{81.0}$ & $\underline{3.7}$ & $2.7$ & $30.8$ & $14.4$ \\
SciTBERT-2022-CI & $26.7$ & $38.9$ & $\underline{42.4}$ & $19.8$ & $22.8$ & $\underline{12.1}$ & $10.4$ & $19.9$ & $19.8$ & $4.6$ & $79.3$ & $\underline{3.7}$ & $2.5$ & $31.2$ & $14.7$ \\
SciTBERT-2023-CI & $\underline{27.0}$ & $40.3$ & $42.3$ & $\underline{20.0}$ & $\underline{23.3}$ & $11.3$ & $10.5$ & $19.8$ & $19.9$ & $4.2$ & $77.8$ & $3.5$ & $2.2$ & $\underline{31.5}$ & $14.6$ \\
SciTBERT-2024-CI & $26.9$ & $39.8$ & $42.3$ & $\underline{20.0}$ & $22.8$ & $11.5$ & $\mathbf{10.7}$ & $\underline{20.6}$ & $20.4$ & $4.9$ & $79.7$ & $3.5$ & $2.4$ & $\underline{31.5}$ & $14.5$ \\
SciTBERT-2025-CI & $26.5$ & $38.2$ & $42.1$ & $19.4$ & $21.7$ & $\underline{12.1}$ & $10.5$ & $19.6$ & $19.6$ & $4.5$ & $79.7$ & $3.6$ & $2.3$ & $30.7$ & $14.2$ \\
\bottomrule
\end{tabular}
\caption{PatRepEval regression and proximity performance, per task. All scores are on a $0$--$100$ scale. Bold denotes best-performing model for the task and underline the second (columnwise).}
\label{tab:pat_rgn_prx}
\end{table*}

\begin{figure}
    \centering
    \includegraphics[width=\linewidth]{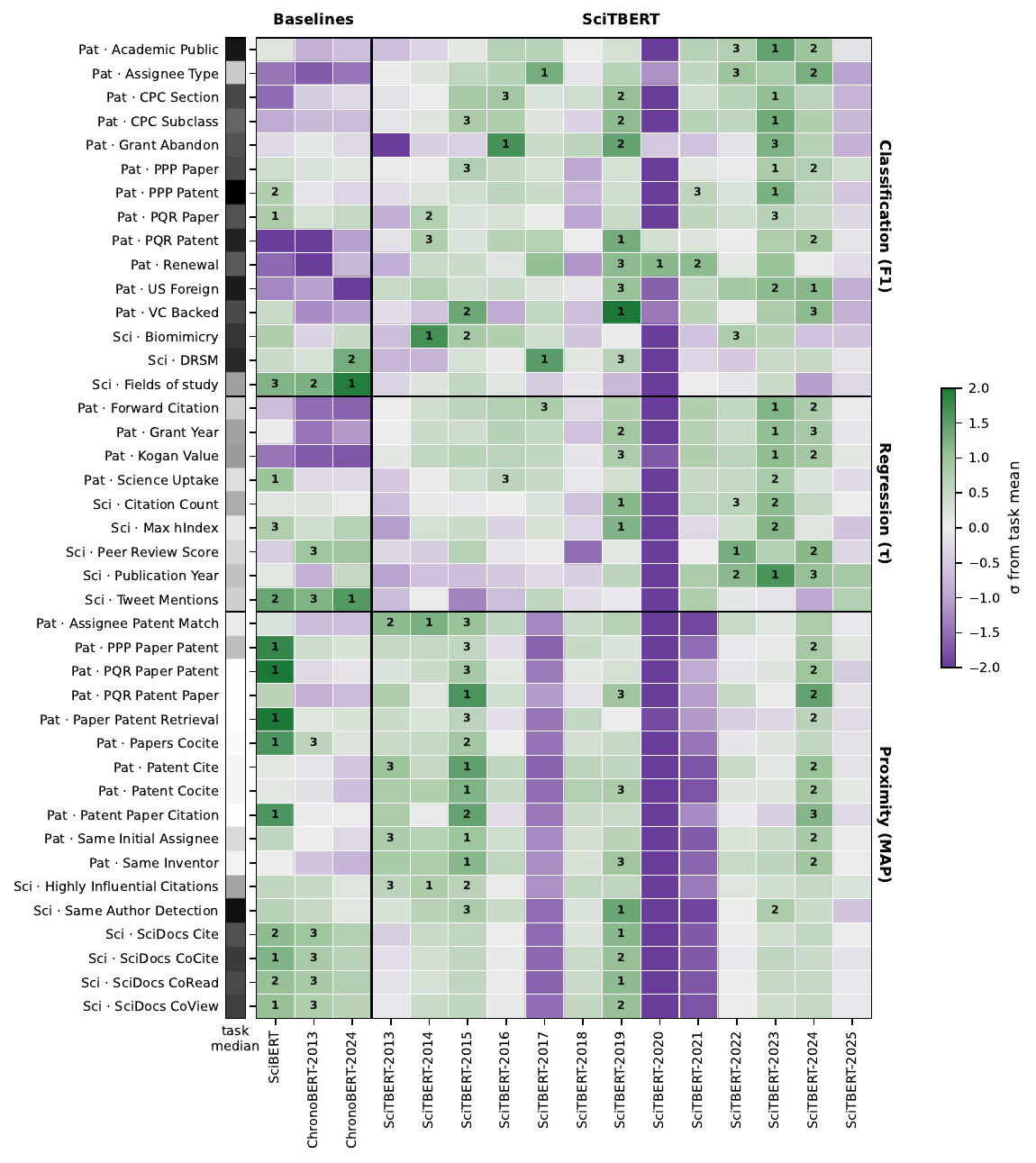}
    \caption{Relative performance of SciTBERT models versus baseline MLM-only scientific and timestamped models across both benchmark tasks. Performance is reported in standard deviations from the task mean performance across the models. The first-, second-, and third-best performing model for the task is denoted with a 1, 2, or 3, respectively.}
    \label{fig:scitbert_vs_mlm}
\end{figure}

\begin{figure}
    \centering
    \includegraphics[width=\linewidth]{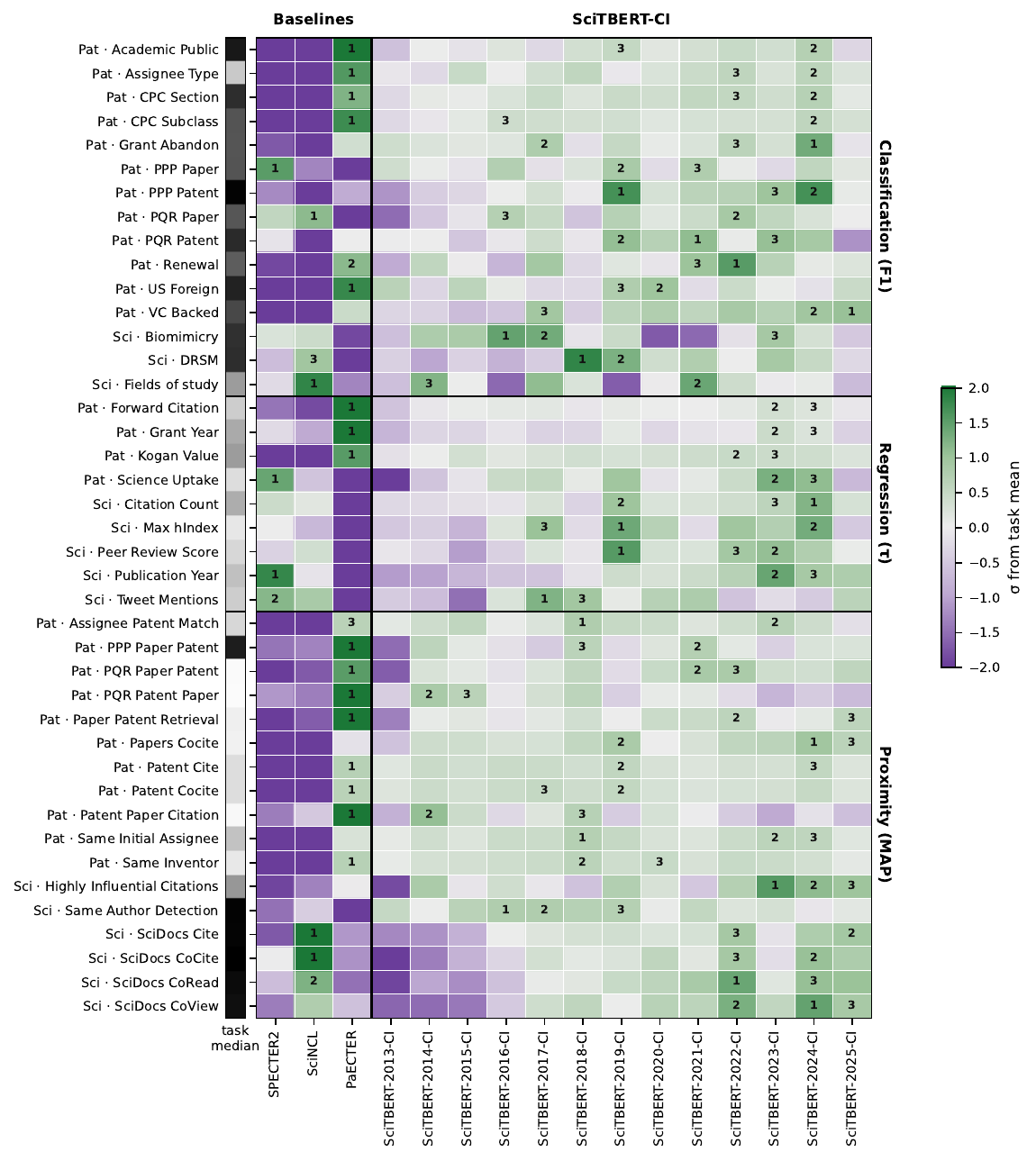}
    \caption{Relative performance of SciTBERT-CI models versus similar contrastively post-trained scientific models. Performance is reported in standard deviations from the task mean performance across the models. The first-, second-, and third-best performing model for the task is denoted with a 1, 2, or 3, respectively.}
    \label{fig:scitbert-ci_vs_ci}
\end{figure}

\end{document}